\documentclass[10pt,twocolumn,letterpaper]{article}

\usepackage{cvpr}              %

\usepackage{graphicx}
\usepackage{amsmath}
\usepackage{amssymb}
\usepackage{booktabs}
\usepackage[accsupp]{axessibility}  %

\usepackage[pagebackref,breaklinks,colorlinks]{hyperref}

\usepackage[capitalize]{cleveref}
\crefname{section}{Sec.}{Secs.}
\Crefname{section}{Section}{Sections}
\Crefname{table}{Table}{Tables}
\crefname{table}{Tab.}{Tabs.}

\makeatletter
\renewcommand\paragraph{\@startsection{paragraph}{4}{\z@}{-0.8mm}{-2.8mm}{\bf\itshape\@adddotafter}}
\makeatother

\usepackage{xcolor}

\newcommand{\mparagraph}[1]{\vspace{1mm}\noindent{\textbf{#1.}\hspace{1mm}}}

\usepackage{import} %
\graphicspath{{figs/}}

\usepackage[ruled,noend]{algorithm2e} %

\SetAlFnt{\small}
\SetAlCapFnt{\small}
\SetAlCapNameFnt{\small}
\SetAlCapHSkip{0pt}
\IncMargin{-\parindent}
\usepackage{ifpdf}
\usepackage{graphicx}
\ifpdf
  
\else
\fi
\usepackage{color}
\definecolor{cyan}{cmyk}{1,0,0,0}
\definecolor{darkgreen}{rgb}{0,0.5,0}
\definecolor{orange}{rgb}{1,0.5,0}
\definecolor{magenta}{cmyk}{0,1,0,0}
\definecolor{darkyellow}{cmyk}{0,0,0.75,0}
\definecolor{gray}{rgb}{0.8,0.8,0.8}

\newenvironment{tightitemize}{
\vspace{-1.2mm}
\begin{itemize}
  \setlength{\itemsep}{1pt}
  \setlength{\parskip}{2pt}
  \setlength{\parsep}{0pt}}{\end{itemize}
\vspace{-1.5mm}
}
\usepackage{amsmath} %

\usepackage[toc,page]{appendix} %
\usepackage{wrapfig} %
\usepackage{comment}
\usepackage{bm}
\usepackage{cuted} %
\usepackage{lipsum} %
\usepackage{mathtools} %
\usepackage{algorithmicx}
\usepackage{algpseudocode}
\usepackage{nicefrac}

\usepackage{gensymb}
\usepackage{float}
\usepackage{soul}
\usepackage{array}
\usepackage{multirow}
\usepackage{hhline}
\usepackage{setspace}
\usepackage{nicefrac}

\algdef{SE}[DOWHILE]{Do}{doWhile}{\algorithmicdo}[1]{\algorithmicwhile\ #1}%

\makeatletter
\renewcommand{\ALG@beginalgorithmic}{\small}
\makeatother

\newcommand{\DELETE}[1]{} %
\newcommand{\IGNORE}[1]{}
\usepackage{datenumber}
\usepackage{calc}
\usepackage[mmddyyyy]{datetime}

\newcounter{datetoday}
\newcounter{diffyears}
\newcounter{diffmonths}
\newcounter{diffdays}

\newcommand{\difftoday}[3]{%
      \setmydatenumber{datetoday}{\the\year}{\the\month}{\the\day}%
      \setmydatenumber{diffdays}{#1}{#2}{#3}%
      \addtocounter{diffdays}{-\thedatetoday}%
      \ifnum\value{diffdays}>0
        \def\diffbefore{}%
        \def\diffafter{left}%
      \else
        \def\diffbefore{}%
        \def\diffafter{ago}%
        \setcounter{diffdays}{-\value{diffdays}}%
      \fi
      \setcounter{diffyears}{\value{diffdays}/365}%
      \setcounter{diffdays}{\value{diffdays}-365*\value{diffyears}}%
      \setcounter{diffmonths}{\value{diffdays}/30}%
      \setcounter{diffdays}{\value{diffdays}-30*\value{diffmonths}}%
      \diffbefore
      \ifnum\value{diffyears}=0
      \else
        \ifnum\value{diffyears}>1
            \thediffyears\space years,
        \else
            \thediffyears\space year,
        \fi
      \fi
      \ifnum\value{diffmonths}=0
      \else
        \ifnum\value{diffmonths}>1
            \thediffmonths\space months
        \else
            \thediffmonths\space month
        \fi
      \fi
      \ifnum\value{diffdays}=0
      \else
        \ifnum\value{diffdays}>1
            \thediffdays\space days
        \else
            \thediffdays\space day
        \fi
      \fi
      \diffafter
}

\graphicspath{{fig/}}

\def\cvprPaperID{00547} %
\def\confName{CVPR}
\def\confYear{2022}

\begin{document}

\title{Polarimetric iToF: Measuring High-Fidelity Depth through Scattering Media} 

\author{Daniel S. Jeon\footnotemark[2]  ~ ~ ~ Andréas Meuleman\footnotemark[2]  ~ ~ ~ Seung-Hwan Baek\footnotemark[1]  ~ ~ ~ 
Min H. Kim\footnotemark[2]\\[5mm]
\footnotemark[2]~ KAIST   ~ ~ ~   ~ ~ ~  \footnotemark[1]~ POSTECH\\
}

\maketitle

\begin{abstract}
\noindent 
Indirect time-of-flight (iToF) imaging allows us to capture dense depth information at a low cost. However, iToF imaging often suffers from multipath interference (MPI) artifacts in the presence of scattering media, resulting in severe depth-accuracy degradation. For instance, iToF cameras cannot measure depth accurately through fog because ToF active illumination scatters back to the sensor before reaching the farther target surface. In this work, we propose a polarimetric iToF imaging method that can capture depth information robustly through scattering media. Our observations on the principle of indirect ToF imaging and polarization of light allow us to formulate a novel computational model of scattering-aware polarimetric phase measurements that enables us to correct MPI errors. We first devise a scattering-aware polarimetric iToF model that can estimate the phase of unpolarized backscattered light. We then combine the optical filtering of polarization and our computational modeling of unpolarized backscattered light via scattering analysis of phase and amplitude. This allows us to tackle the MPI problem by estimating the scattering energy through the participating media. We validate our method on an experimental setup using a customized off-the-shelf iToF camera. Our method outperforms baseline methods by a significant margin by means of our scattering model and polarimetric phase measurements.
\end{abstract}

\section{Introduction}
\label{sec:intro}
\noindent Time-of-Flight (ToF) imaging is the cornerstone of modern 3D imaging technology that has received great attention across diverse fields, including computer graphics and vision.
Its notable applications include autonomous driving, 3D motion capture, digital-human reconstruction, human-computer interfaces, robotics, etc.
Modern ToF cameras can be broadly categorized into direct and indirect systems.
Direct ToF measures the round-trip time of photons emitted from an illumination source until they travel back to the ToF detector.
Indirect ToF, referred to as amplitude-modulated continuous-wave ToF, utilizes a temporally modulated illumination source and \emph{computationally} estimates the round-trip time of photons from modulation phase changes~\cite{lange2001solid}.
The indirect acquisition principle lowers the system-building cost by departing from the necessity of the \emph{picosecond-accurate} illumination,  detector, and synchronization module used in direct ToF.
Furthermore, indirect ToF achieves low-cost instant 3D imaging of the entire field of view with flood-fill illumination. %
As a result, indirect ToF cameras have achieved remarkable success in commercial markets, e.g.,  Microsoft Azure Kinect and PMD sensors.

\begin{figure}
\vspace{-3mm}%
\centering\includegraphics[width=0.95\linewidth]{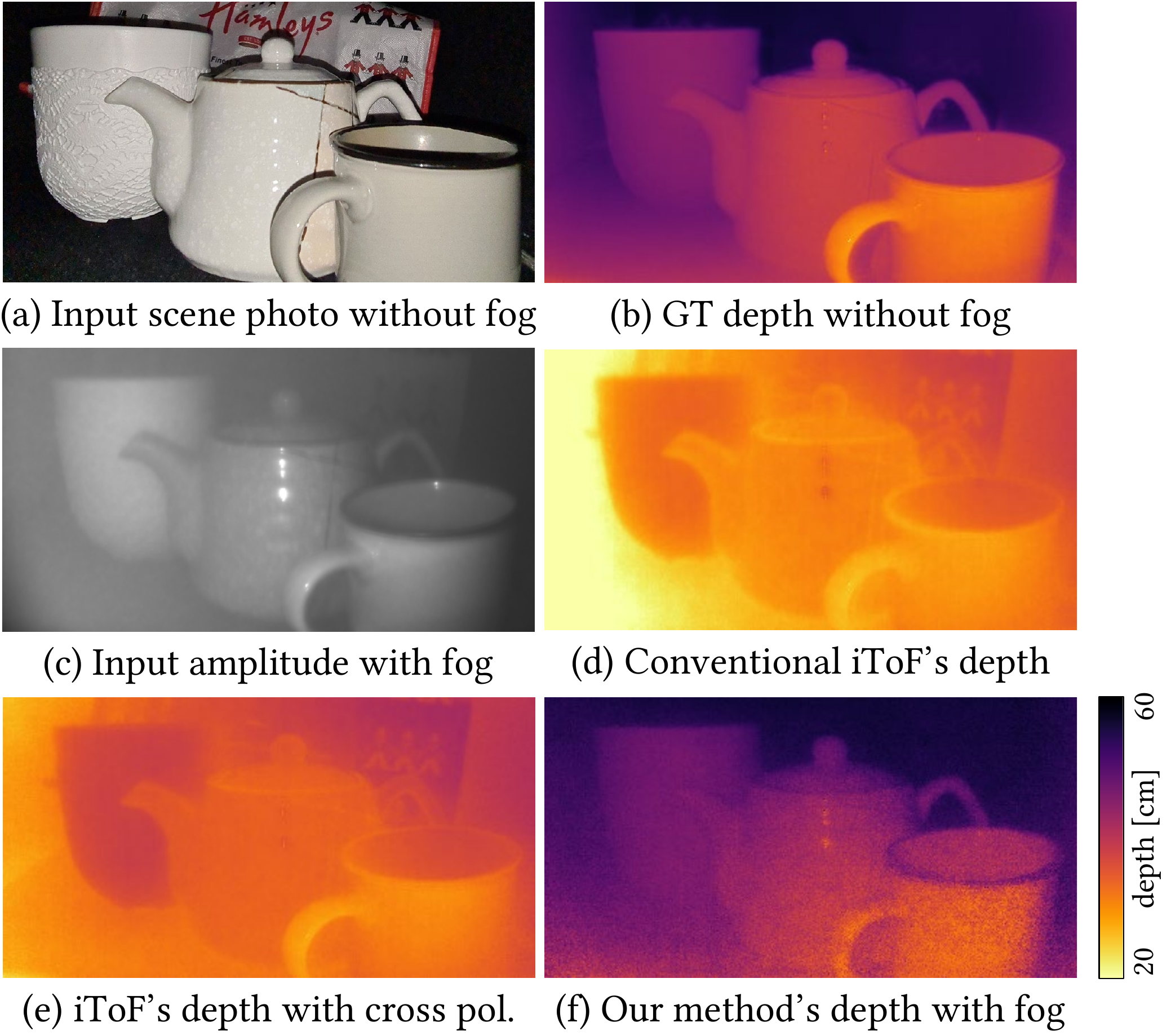}\vspace{-3mm}%
		\caption{\label{fig:teaser}%
		We introduce a polarimetric iToF imaging method that can estimate depth robustly through scattering media. %
		(a)~A photograph of the input scene without fog. 
		(b)~Ground-truth depth measure without fog.
		(c)~Input iToF amplitude map captured with fog.
		(d)~Depth estimated by a conventional iToF camera with fog. 
		(e)~Depth improved by na\"ive cross-polarization filtering.
		(f)~Our iToF depth measurement result is fairly close to the GT depth. 
	}\vspace{-3mm}%
\end{figure}

However, it is also the \emph{indirect}-imaging scheme that poses critical limitations on robust 3D imaging.
One of the notable resulting challenges is multi-path interference (MPI).
Light emitted from the ToF illumination module travels through a scene and reaches the ToF sensor.
During light transport, some photons interact with only one scene point via direct reflection, thus providing accurate depth information of that point.
However, other photons undergo multiple reflections on different scene points
because of indirect reflection.
If a pixel on the ToF sensor receives a mixture of direct and indirect photons, the measured phase shift does not correspond to the analytical phase shift of the target scene depth anymore.
Thus, it degrades the accuracy of the reconstructed depth.

The MPI problem becomes more severe in the presence of scattering media such as fog (Figure~\ref{fig:teaser}(a) for example) because light photons experience numerous indirect reflections with the scattering particles.
In this case, the scattered light energy often exceeds that of light interacting with a target scene point, resulting in extremely inaccurate scene depth estimation as shown in Figure~\ref{fig:teaser}(c),
i.e., the measured distance through fog tends to be closer than the actual distance.
This acts as a critical hurdle for indirect ToF cameras to be deployed in the wild,
e.g., fire-rescuing robots, autonomous driving under fog, and underwater navigation. %

In this paper, we propose a polarimetric iToF imaging method robust to scattering environments. 
Our key idea is to revisit the polarization of light and the scattering theory about intensity attenuation and depolarization.
Our method allows for accurate scene depth estimation even in the presence of severe scattering media, as shown in Figure~\ref{fig:teaser}(d).

We leverage the polarization property of light that the backscattered light from scattering particles better maintains the polarization state of the emitted photons than the light that travels farther to a surface~\cite{chenault2000polarization}.
We first configure the orthogonal polarization modulation of ToF illumination and detection to initially filter out the polarized backscattered light optically.
While existing methods~\cite{fade2014long, zeng2018visible, van2015detection, gilbert1967improvement} also demonstrate the effectiveness of this cross-polarization setup,
one critical problem of cross-polarization setup is that the assumption on the polarized state of backscattered light does not hold in practice because backscattered light undergoes a change of polarization throughout scattering events toward an unpolarized state~\cite{wang2020analyzing}.
This results in limited depth accuracy.

To handle this, we devise a computational method that can eliminate the remaining unpolarized backscattered light based on the indirect ToF's signal representation: phase and amplitude.
First, we estimate the phase of unpolarized backscattered light by revisiting the scattering model of intensity attenuation and depolarization~\cite{shashar2004transmission}.
Second, the amplitude of unpolarized backscattered light is estimated based on the observation that the amplitude-offset ratio is consistent for non-scattered light.
Then, our method subtracts the unpolarized backscattered light from the initial cross-polarization measurements, resulting in the estimates of scattering-free indirect ToF measurements.
Our polarimetric iToF imaging method can enhance depth accuracy significantly, outperforming existing baselines for depth estimation through scattering media, as shown in Figure~\ref{fig:teaser}(d).

In summary, our contributions are:
\begin{tightitemize}
	\item A scattering-aware polarimetric phasor model specifically designed for polarimetric iToF imaging, based on the scattering theory of light intensity attenuation and depolarization.
	\item An efficient scattering phasor optimization that can estimate the phase of unpolarized backscattered light via scattering analysis of phase and amplitude in iToF.
\end{tightitemize}

\section{Related Work}
\label{sec:relatedwork}

\mparagraph{Multi-path interference}
Indirect ToF cameras measure the round-trip time of light emitted from an amplitude-modulated illumination source until it travels through a scene and is captured by a ToF detector.
While being a practical depth-imaging technology, indirect ToF imaging suffers from MPI artifacts.
As we capture the sum of directly-reflected light from a scene and indirectly-reflected light through multiple reflections, iToF often results in distorted phase measurements. %
The MPI problem can be mitigated by extracting direct-only reflection from such intergraded measurements. %
One effective approach is to capture iToF measurements with multiple modulation frequencies~\cite{freedman2014sra, patil2020depth,jimenez2014modeling,fuchs2010multipath,gupta2015phasor}.
Another direction is to utilize the data-driven depth prior of natural scenes to estimate the direct-only reflection from the mixture of direct/indirect reflections~\cite{guo2018tackling,su2018deep,agresti2018deep,marco2017deeptof}.
While these methods can deal with scenes containing second-bounce reflections of light, they often fail to handle more extreme scenarios, such as scenes with scattering media, where scattering events make the number of reflections substantially higher than two.

Using an analytical scattering model of intensity attenuation is an effective solution in ToF imaging~\cite{mure2007optimized}.
Fujimura et al.~\cite{fujimura2020simultaneous} extend the scattering-model approach by utilizing segmented background pixels that are only contributed by backscattered light without any light from scene reflection.
However, capturing natural scenes often violates this assumption.
One can overcome this background-dependency problem by using relatively short-pulse ToF imaging~\cite{kijima2021time} at an increased cost for building a picosecond-accurate synchronized ToF camera like direct TOF.

\mparagraph{Polarization and scattering}
Polarization of light describes how its electric field oscillates in space~\cite{kerker2013scattering}. As a wave property of light, polarization has been extensively utilized for many graphics and vision problems including shape from polarization~\cite{baek2018simultaneous,fukao2021polarimetric}, appearance from polarization~\cite{baek2020image}, light transport~\cite{baek2021polarimetric}, direct ToF imaging~\cite{baek2022all}, and reflection removal~\cite{nayar1993removal,lei2020polarized}.

Most relevant to us, polarization helps us see through scattering media by optically filtering out the backscattered light and only capturing light that has interacted with the target surface using polarization.

As a common practice for achieving this goal, one can use two linear polarizers at a perpendicular configuration in front of an illumination module and a camera~\cite{zeng2018visible}.
This setup, so-called cross-polarization, optically rejects  backscattered light, which tends to maintain the polarization state of illumination, thus filtered out by the perpendicular polarizer on the detector~\cite{konnen1985polarized}.
In contrast, light that has traveled to a target surface mostly loses the polarization state of the original illumination, therefore, can be detected by the camera passing through the perpendicular polarizer.

An extension to cross-polarization imaging is polarization-difference imaging (PDI) which takes an additional image with a parallel orientation of the two polarizers instead of the perpendicular configuration~\cite{rowe1995polarization,nan2009linear}.
Subtracting the cross-polarization measurements from the parallel-polarization measurements helps us estimate the backscattered light~\cite{treibitz2008active}.
PDI offers a better imaging capability in the presence of scattering media than cross-polarization imaging, which can be further improved using the segmented MPI-free background pixels~\cite{zhang2022time, zhao2022polarization}.
However, they still suffer from limited depth-imaging capability because of the unmet assumption on the spatially-uniform polarization state of backscattered light. %
Real-world scattered light exhibits spatially-varying polarization states~\cite{fujimura2020simultaneous} especially when an active illumination is used as in ToF imaging. %
Our polarimetric iToF method does not make such assumptions and thus enables accurate 3D imaging even under severe scattering media.

\section{Background}
\label{sec:background}
\begin{figure}[tp]
\vspace{-3mm}%
	\centering	
	\includegraphics[width=0.75\linewidth]{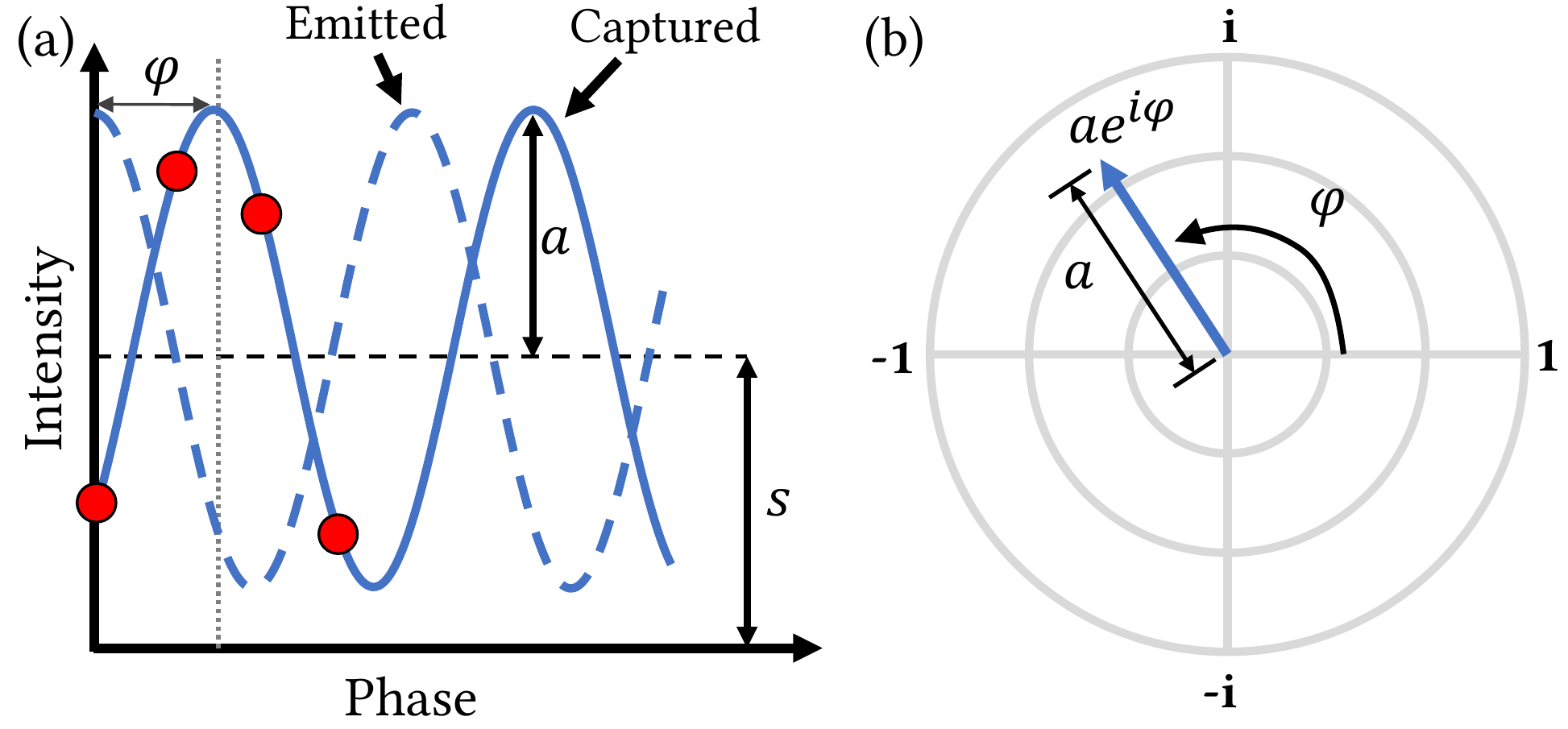}%
	\vspace{-4mm}%
	\caption[]{\label{fig:tof_phasor_representation}%
    Phasor representation of iToF imaging. (a)~Amplitude-modulated light returns to the iToF sensor after traveling a scene. Phasor representation includes amplitude $a$, phase $\varphi$, and offset $s$ to describe the continuous signal. Indirect ToF imaging measures four samples at different phases indicated by the red dots.  (b)~We visualize the amplitude and phase of the continuous signal in the polar coordinates, where the length and direction of a vector are used to represent amplitude and phase. %
	}%
	\vspace{-5mm}
\end{figure}

\noindent Indirect ToF cameras emit and capture continuously amplitude-modulated light, which can be characterized with three parameters, called phasor representation~\cite{gupta2015phasor}: amplitude~$a$, phase~$\varphi$, and offset $s$.
Figure~\ref{fig:tof_phasor_representation} shows a polar-coordinate visualization of the phasor representation, where the length and angle of the vector correspond to the amplitude and phase of the signal. %

In iToF imaging, we obtain the phasor representation by capturing multiple samples of the returning light at different phases.
A common choice is to use four-phase samples at
$\phi=\{0^\circ, 45^\circ, 90^\circ, 135^\circ\}$, resulting in the sampled intensities of light as $\{I_\phi\}$.
Once measured, the phasor representation is expressed as:
\vspace{-3mm}
\begin{align}
	\label{eq:recon_phasor}
&\textrm{Amplitude: } 	& a=&\frac{1}{2}\sqrt{(I_{135^\circ} - I_{45^\circ})^2 + (I_{90^\circ} - I_{0^\circ})^2}, \nonumber \\
&\textrm{Phase: }	& \varphi=&\text{arctan2} \left( {{I_{135^\circ} - I_{45^\circ}},{I_{90^\circ} - I_{0^\circ}}} \right), \nonumber \\
&\textrm{Offset: }    & s=&\left(I_{0^\circ} + I_{45^\circ} + I_{90^\circ} + I_{135^\circ}\right) / 4. 
\end{align}

\section{Method}
\label{sec:ourmethod}
\noindent
We first model how the backscattered light distorts the true phasor of a scene point in consideration of light polarization. %
Our analytical model then enables us to remove the undesired phasor distortion from polarimetric iToF measurements for accurate 3D imaging.

\begin{figure}[tp]
	\vspace{-3mm}%
	\centering	
	\includegraphics[width=1.0\linewidth]{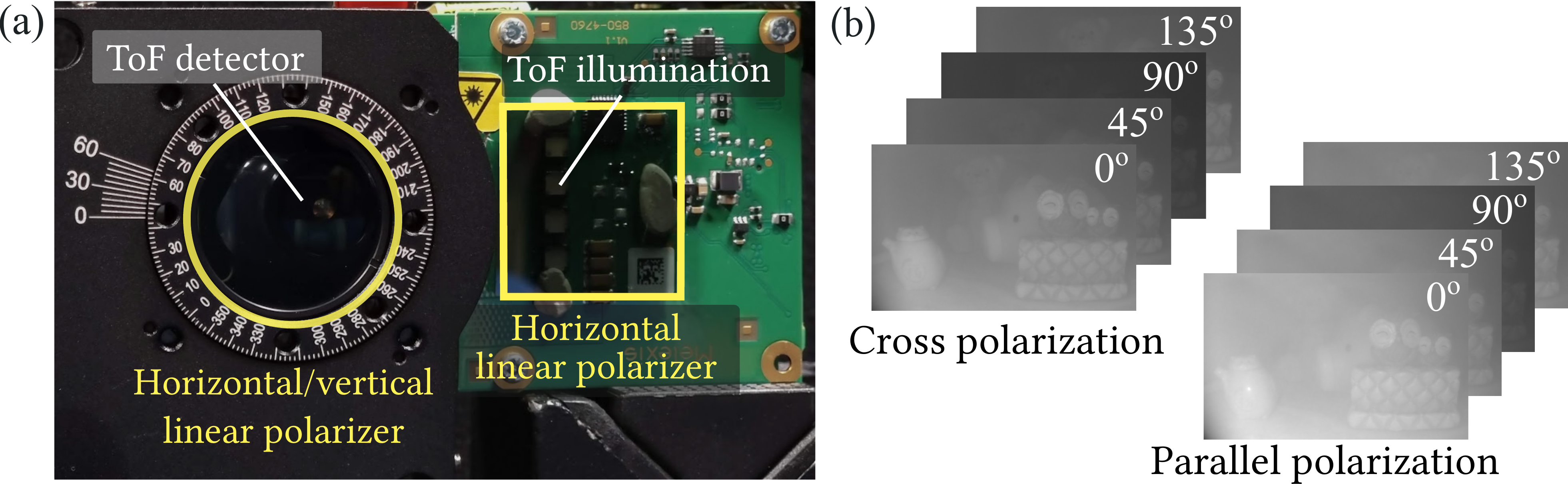}%
	\vspace{-3mm}%
	\caption[]{\label{fig:system_configuration}%
(a) Our polarimetric iToF imaging setup.
(b) We capture four-tap phase samples with cross and parallel polarization, respectively, as input.
	}%
	\vspace{-6mm}
\end{figure}

\mparagraph{Imaging setup}
We equip an off-the-shelf ToF module with two linear polarizers: one for the light source and another for the detector.
While the linear polarizer on the light source is set to the horizontal orientation, the linear polarizer in front of the detector is mounted on a motorized rotation stage to provide two orthogonal angles.
As input, we use two sets of four-tap $\phi$ phase measurements of the parallel and perpendicular orientations of the detector's linear polarizer, respectively.
See Figure~\ref{fig:system_configuration}.

\subsection{Input}
\label{sec:formation}
\noindent We describe the captured light $I_\phi$ from the customized iToF camera as the sum of scattered light $S_\phi$ and target light $T_\phi$ that has interacted with scene objects:
\vspace{-2mm}
\begin{equation}\label{eq:light}
  I_\phi=S_\phi+T_\phi=\left(S_\phi^{u} + S_\phi^{p}\right) + \left(T_\phi^{u} + T_\phi^{p}\right),
\end{equation}
where $\{S/T\}_\phi^{u}$ and $\{S/T\}_\phi^{p}$ are the unpolarized and polarized components for each case. %

\mparagraph{Unpolarized input}
When a perpendicular orientation of the polarizers is set, i.e., \emph{cross-polarization} configuration, the measurement is not affected by the light polarized in the same direction as the illumination, resulting in the following image formation:
{$  I_\phi^{\bot}=\frac{1}{2}S_\phi^u + \frac{1}{2}T_\phi^u,$}
where $I_\phi^{\bot}$ is the captured light intensity of cross-polarization.

The unpolarized backscattered light $S_\phi^u$ is often ignored ($S_\phi^u\approx0$) in conventional cross-polarization imaging, enabling a straightforward computation of the target-only signal $T_\phi^u \approx I_\phi^{\bot}$. %
However, this assumption does not practically hold.
In fact, only a fraction of backscattered light from a very near distance meets this requirement.
Light photons that undergo multiple scattering events lose the polarization state of the original illumination, turning into unpolarized light $S_\phi^u$ ~\cite{jarry1998coherence, lewis1999backscattering}.
The unmet assumption results in inaccurate 3D imaging under scattering media.

\mparagraph{Polarized input}
To avoid the previous assumption, we capture another polarimetric phase image with the parallel orientation of the illumination-detection linear polarizers by rotating the linear polarizer on the detector side.
Note that the parallel-polarization configuration does not reject the polarized component $\{S/T\}_\phi^p$ as in the cross-polarization setup because the returning light with the same polarization state as the illumination still passes through the parallel-oriented polarizer on the detector.
Hence, we model the captured light intensity $I_\phi^{\parallel}$ as
{$  I_\phi^{\parallel}=S_\phi^p + \frac{1}{2}S_\phi^u + T_\phi^ p + \frac{1}{2}T_\phi^u.$}

\begin{figure}[tp]
\vspace{-3mm}%
	\centering	
	\includegraphics[width=0.9\linewidth]{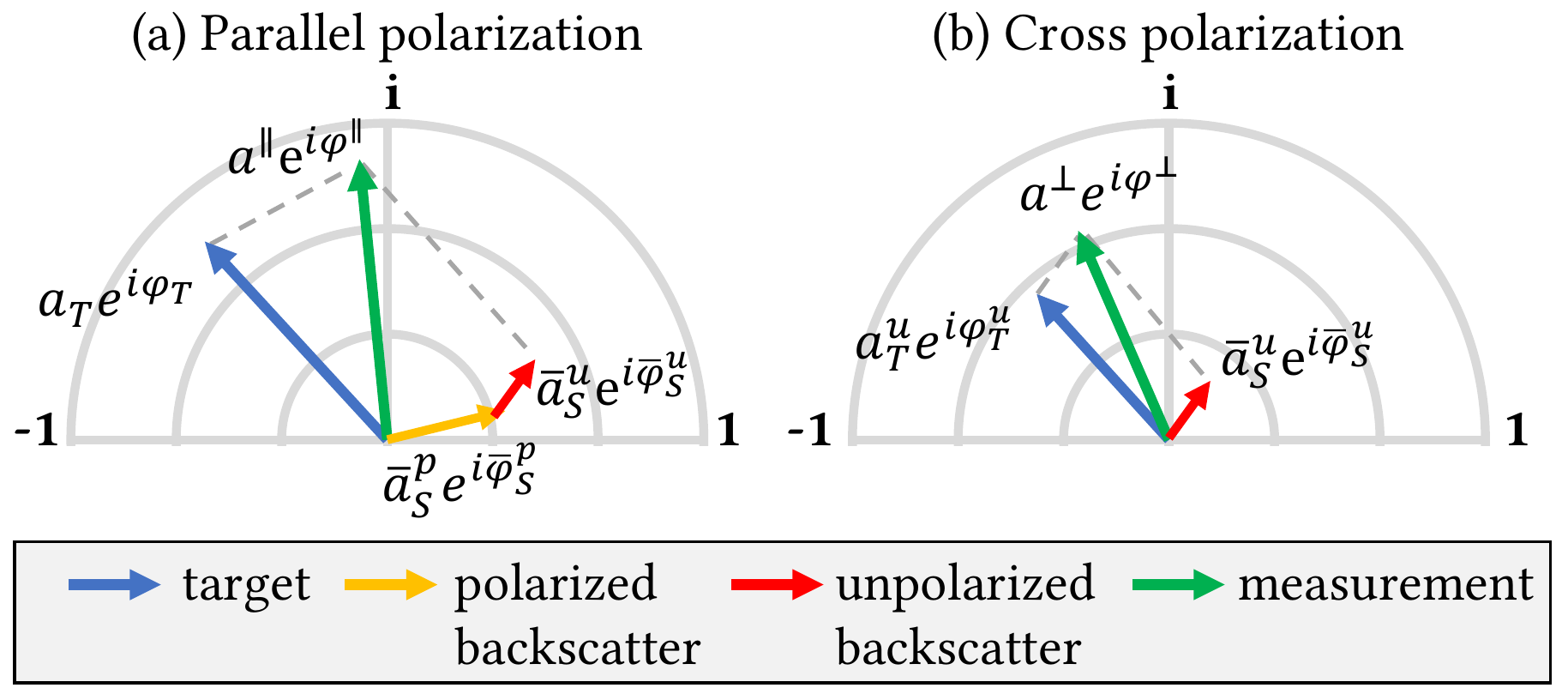}%
	\vspace{-3mm}%
	\caption[]{\label{fig:fog_phasor_representation}%
    Phasor of the target, polarized backscatter, unpolarized backscatter, and measurements for (a) the parallel-polarization setting and (b) the cross-polarization setting.
    The phase of the target light stays the same and only the amplitude is different between the two captures. %
    In contrast, both phase and amplitude change for the total backscattered light due to the contribution of polarized backscatter. Note that the polarized backscatter disappears in the cross-polarization setting.
	}%
	\vspace{-6mm}
\end{figure}

\subsection{Phasor Model}
\label{sec:overview}
\noindent
Figure~\ref{fig:fog_phasor_representation} depicts our phasor image formation model.
With the ultimate goal of estimating the \emph{phase} information of the \emph{target} point (blue arrows) from given (a) parallel-polarized and (b) cross-polarized phasor measurements (green arrows),
we want to estimate the phasor information of the backscattered light first.
To this end, our method should know two additional phasor representations: (1)~the polarized backscattered light (yellow arrow) and (2) the unpolarized backscattered light (red arrow).

\mparagraph{Phasor of polarized scattering}
The first one is the phasor representation of the \emph{polarized} backscattered light (yellow arrow) which can be easily obtained following the principle of PDI~\cite{rowe1995polarization}.
We subtract the cross-polarization measurement $I_\phi^{\bot}$ from the parallel measurement~$I_\phi^{\parallel}$:
 $ I_\phi^{-}=I_\phi^{\parallel}-I_\phi^{\bot}=S_\phi^p  + T_\phi^ p \approx S_\phi^p. $
The target light reflected from a scene point is unlikely to have the same polarization state as the original illumination due to the numerous scattering events during its light transport.
Since most light from the target surfaces is the diffuse reflection and has interacted with many scattering particles, it is unpolarized: $T_\phi^ p \approx 0$.

As a result, we can obtain the phasor information of \emph{polarized backscattered light} $S_\phi^p$ from the PDI measurements $I_\phi^{-}$ using using Equation~\eqref{eq:recon_phasor}, yielding phase shift $\overline{\varphi}_{S}^p$, amplitude $\overline{a}_{S}^p$, and offset $\overline{s}_{S}^p$.

\mparagraph{Phasor of unpolarized scattering}
Second, given the phasor information of the polarized backscattered light (orange arrow), we need to know the phasor representation of \emph{\underline{unpolarized backscattered light}}  (red arrow in Figure~\ref{fig:fog_phasor_representation}): phase shift $\overline{\varphi}_{S}^u$, amplitude $\overline{a}_{S}^u$, and offset $\overline{s}_{S}^u$ to obtain the target phase information (blue arrow). 
The following section provides details of our solution.

\subsection{Phasor of Unpolarized Backscattered Light}
\label{sec:phasor-all}
\noindent We estimate the phasor of unpolarized backscattered light.
Phasor consists of phase and amplitude, which are optimized in two folds.
We first estimate the \emph{scattering decay parameter} of backscattered light according to the volumetric scattering theory.
We then obtain \emph{phase} and \emph{amplitude} from unpolarized backscattered light.

\subsubsection{Scattering Decay}
\label{sec:scattering-decay}

To estimate the phase of the unpolarized backscattered light, we obtain the integrated phase shift of \emph{polarized} multiple backscattered light ${\overline{\varphi}_S^p}$ based on volumetric integration over the entire range of phase shift~$\varphi$:
\begin{equation}
	\label{eq:phase_of_scattering_polarized_from_model_phi}
    {\overline{\varphi}_S^p} = \frac{\int_{\varphi_0}^\infty  {\varphi {a^p_S}( \varphi )  }d\varphi}{\int_{\varphi_0}^\infty  {{a^p_S}( \varphi  ) } d\varphi},
\end{equation}
where ${a^p_S}( \varphi )$ is a phase-conditioned amplitude function, and $\varphi_0$ is the phase shift corresponding to the nearest travel distance of ToF light.
We can interpret Equation~\eqref{eq:phase_of_scattering_polarized_from_model_phi} as a weighted average of phase shift $\varphi$ with the weight of the amplitude function ${a^p_S}( \varphi )$.

\begin{figure}[tp]
	\centering	
	\vspace{-4mm}	
    \def\svgwidth{\linewidth}\footnotesize
    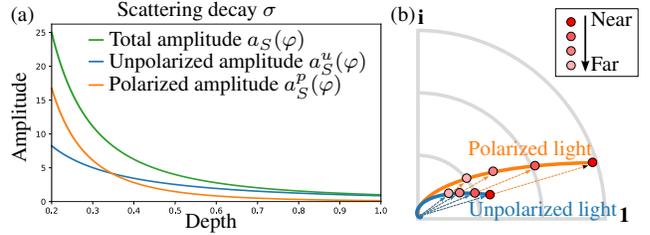
	\vspace{-7mm}%
	\caption[]{\label{fig:our_method_polarization_fraction}%
    (a) Exponentially-decaying scattering model describes the amplitude changes of polarized and unpolarized backscattered light with respect to depth. For a far depth, the contribution of the unpolarized light becomes larger than that of the polarized light. %
    (b) Phasor representation of unpolarized light and polarized light along depth, here only shown the first quadrant. The phasor is the sum of continuously varying phasors along depth.
	}%
	\vspace{-5mm}
\end{figure}

As shown in Figure~\ref{fig:our_method_polarization_fraction}, we formulate the amplitude function ${a^p_S}( \varphi )$ based on the exponential decay of amplitude and degree-of-polarization (DoP) in scattering media~\cite{shashar2004transmission, wilkie2004analytical, sankaran2002comparative} as follows:
\begin{equation}
	\label{eq:amplitude_of_scattering_polarized}
{a^p_S}\left( \varphi  \right) \propto {\frac{1}{\varphi^2}} \underbrace{\rm{exp}({ - {\sigma_i}\varphi }}_{\mathrm{intensity}}) \underbrace{{\rm{exp}({ - {\sigma_p}\varphi }})}_{\mathrm{DoP}},
\end{equation}
where $\sigma_i$ and $\sigma_p$ are the extinction coefficients of \emph{intensity} and \emph{DoP} attenuation.
The first term $\frac{1}{\varphi^2}$ comes from the inverse-square law of emitted light into a scene. The second and third terms describe the exponential decay of \emph{polarized} light amplitude and DoP.

After substituting \(a_S^p\) in Equation~\eqref{eq:phase_of_scattering_polarized_from_model_phi} with Equation~\eqref{eq:amplitude_of_scattering_polarized}, we integrate the numerator and the denominator of Equation~\eqref{eq:phase_of_scattering_polarized_from_model_phi}  using the exponential integral formula:
\begin{align}
	\label{eq:phase_of_scattering_polarized_from_model_solution}
    \overline{\varphi}_S^p = f(\sigma) = \frac{  \rm{Ei}\left( { \sigma {\varphi_0}} \right)}{-\sigma \rm{Ei}\left( { \sigma {\varphi_0}} \right) + \frac{1}{\varphi _0}\rm{exp}({- \sigma {\varphi _0}})} ,
\end{align}
where $\sigma$ is the total decay rate, the sum of intensity $\sigma_i$ and DOP $\sigma_p$, and $\rm{Ei}(\cdot)$ is the exponential-integral function. The initial phase $\varphi_0$ can be obtained by geometric calibration of the system.
As a result, we have established our model $f(\cdot)$ on the phase shift of polarized backscatter $\overline \varphi _S^p$ as a function of the decay parameter~$\sigma$.

We estimate the best scattering decay parameter $\sigma$ that produces the prediction of $f(\sigma)$ most similar to the experimental data~$\overline \varphi _S^p$ obtained in Section~\ref{sec:overview}:
\begin{equation}\label{eq:decay_optimization}
  \underset{\sigma}{\mathrm{minimize}}\, \| \overline \varphi _S^p - f(\sigma) \|_2^2.
\end{equation}
We solve this using the Adam gradient-descent optimization.
We obtain the decay rate $\sigma$ as a median value from its per-pixel estimates. %

\subsubsection{Phase Estimation}
\label{sec:phase-estimation}
Once the decay parameter $\sigma$ is estimated, we turn to estimate the phase of \emph{unpolarized} backscattered light.
To this end, we develop an \emph{unpolarized} version of Equations~\eqref{eq:phase_of_scattering_polarized_from_model_phi} and \eqref{eq:amplitude_of_scattering_polarized}.
The integrated phase shift ${\overline\varphi_S^u}$ of unpolarized backscattered light is defined as
\begin{equation}
	\label{eq:phase_of_scattering_unpolarized_from_model}
    {\overline\varphi_S^u} = \frac{\int_{\varphi_0}^\infty  {\varphi {a_S^u}\left( \varphi  \right)d\varphi  }}{\int_{\varphi_0}^\infty  {{a_S^u}\left( \varphi  \right)d\varphi } },
\end{equation}
where we use the amplitude of unpolarized backscattered light ${a_S^u}(\varphi)$ as a weight.
We then define the amplitude function by considering the unpolarized ratio as
\begin{equation}
	\label{eq:amplitude_of_scattering_unpolarized}
{a^u_S}\left( \varphi  \right) \propto {\frac{1}{\varphi^2}} \underbrace{\rm{exp}({ - {\sigma_i}\varphi })}_{\mathrm{intensity}} \underbrace{(1-{\rm{exp}({ - {\sigma_p}\varphi })})}_{\mathrm{1-DoP}}.
\end{equation}
Note that the third term attenuates the amplitude with the ratio of unpolarized light.

Similarly with the polarized case, we rewrite Equation~\eqref{eq:phase_of_scattering_unpolarized_from_model} by substituting ${a^u_S}\left( \varphi  \right)$ with Equation~\eqref{eq:amplitude_of_scattering_unpolarized}:
\begin{align}\label{eq:phase_of_scattering_unpolarized_from_model_solution}
    &{\overline \varphi_S^u} =  \\
	&\resizebox{1.0\linewidth}{!}{
	\mbox{\fontsize{10}{12}\selectfont $
\frac{{ \rm{Ei}\left( { {\sigma _i}{\varphi_0}} \right) - \rm{Ei}\left( { \sigma {\varphi_0}} \right)}}{{{-\sigma _i}\rm{Ei}\left( {  {\sigma _i}{\varphi_0}} \right) + \sigma \rm{Ei}\left( { \sigma {\varphi_0}} \right) + \frac{1}{\varphi_0}{{\rm{exp}({ - {\sigma _i}{\varphi_0}})} - \frac{1}{\varphi_0}{\rm{exp}({ - \sigma {\varphi_0}})}}}} .
$ } } \nonumber
\end{align}
Since we have estimated the sum of decay rates $\sigma$ from Equation~\eqref{eq:decay_optimization} already,
we can exclude $\sigma$ from the function parameter.

Lastly, Equation~\eqref{eq:phase_of_scattering_unpolarized_from_model_solution} is the analytical model of the phase of unpolarized backscattered light and allows us to compute the phase distortion if the decay rate of intensity $\sigma_i$ is known.
To this end, we utilize the previously estimated integrated decay rate $\sigma$ from Equation~\eqref{eq:decay_optimization}.
Both intensity and DoP decrease exponentially under the same scattering media with respect to travel distance.
Thus, the total decay rate $\sigma=\sigma_i+\sigma_p$ can be related to the intensity decay rate as $\sigma_i = \alpha \sigma$, where $\alpha$ is a global scalar.
We calibrate the scalar $\alpha$ using a fog chamber, the details of which are included in the supplemental document.
With the calibrated scalar $\alpha$ and the previously estimated $\sigma$, we compute $\sigma_i$ which is then used to compute the phase of unpolarized backscattered light using Equation~\eqref{eq:phase_of_scattering_unpolarized_from_model_solution}.

\subsubsection{Amplitude Estimation}
\label{sec:amplitude-estimation}
\mparagraph{Amplitude-to-offset ratio}
Before estimating the scattered amplitude of unpolarized light, we first define the amplitude-to-offset ratio.
Suppose a scene has no scattering media and interreflection.
An indirect ToF camera then captures direct reflection $T$ only. %
In this scenario, the ratio of phasor amplitude and offset is constant:
\begin{equation}\label{eq:consistency}
  {a_T}/{s_T} = k_0,
\end{equation}
where $k_0$ is the amplitude-to-offset ratio that only depends on the power range of the ToF illumination module.
Note that the ratio is independent of scene reflectance.
We calibrate $k_0$, which is 0.71 in our setup, by capturing a reference target in a darkroom.

\mparagraph{Amplitude modeling}
Our main goal is to estimate the unknown scattered amplitude ${\overline a^u_S}$ of unpolarized light from the given information: (a) the calibrated amplitude-offset constant $k_0$, (b) the \emph{analytical} phasor representation of the polarized/unpolarized/PDI measurements
 and (c) our phase estimate of unpolarized backscattering $\overline \varphi_S^u$ in Section~\ref{sec:phase-estimation}.
We first formulate the phasor of unpolarized target light as:
\begin{align}
    a_T^u & =  \left\|  {a^\bot {\rm{exp}({i \varphi^\bot}) }} - {\overline a_S^u {\rm{exp}({i \overline \varphi_S^u}) }} \right\|, \label{eq:unpolarized_light_a} \\
  s_T^u & =  s^\bot - \overline s_S^u,\label{eq:unpolarized_light_b}
\end{align}
where $a^\bot$, $\varphi^\bot$, and $s^\bot$ are the phasor of the cross-polarization measurements obtained by Equation~\eqref{eq:recon_phasor}.
Note that we aim to estimate the amplitude~$\overline a_S^u$ in this equation.
$\overline a_S^u $, $\overline \varphi_S^u$, and $\overline s_S^u$ are the \emph{integrated} amplitude,
phase, and offset of unpolarized multiple backscattered light, which will be modeled in the following.

\mparagraph{Amplitude modeling with amplitude-to-offset ratio}
We now apply the constant amplitude-to-offset ratio of Equation~\eqref{eq:consistency} to the unpolarized \emph{target} light and the unpolarized \emph{backscattered} with a specific phase shift $\varphi$:
\begin{align}
{a_T^u}/{s_T^u} =  k_0, \label{eq:consistency_unpolarized} \\
  a_S^u (\varphi)/s^u_S(\varphi)=k_0. \label{eq:offset_unpolarized_no_MPI}
\end{align}

For the \emph{integrated} backscattered light over the phase shift $\varphi$, the constant amplitude-to-offset ratio does not hold anymore.
In fact, the ratio $\overline k^u_S$ decreases lower than $k_0$, depending on the thickness of the scattering media and interreflection:
\begin{align}\label{eq:offset_unpolarized}
&\frac{\overline{a}_S^u}{\overline{s}_S^u} =   \overline k^u_S\\
&\resizebox{0.9\linewidth}{!}{
	\mbox{\fontsize{10}{12}\selectfont $
=  \frac{{\left\| {\int_{{\varphi _0}}^\infty  a_S^u (\varphi) \rm{exp}({i \varphi}) } d\varphi \right\|} }{{\int_{{\varphi _0}}^\infty s^u_S(\varphi) d\varphi }}
 =  k_0 \frac{{\left\| {\int_{{\varphi _0}}^\infty  a_S^u (\varphi) \rm{exp}({i \varphi}) } d\varphi \right\|} }{{\int_{{\varphi _0}}^\infty a^u_S(\varphi) d\varphi }},
$ } } %
\nonumber
\end{align}
where $\overline k^u_S$ can be rewritten as a function of $a^u_S(\varphi)$ using Equation~\eqref{eq:offset_unpolarized_no_MPI} as shown on the right-hand-side of Equation~\eqref{eq:offset_unpolarized}.
Note that $k_s$ is the ratio of amplitude to offset, and $k_s$ is smaller than $k_0$ because the amplitude of the integrated phasor is smaller than the integration of the amplitudes themselves.

Lastly, since the target amplitude model $a_T^u$ in Equation~\eqref{eq:unpolarized_light_a} includes the \emph{amplitude of unpolarized backscattered light $\overline a_S^u$} that we want to find out,
we first combine Equations~\eqref{eq:unpolarized_light_a} and~\eqref{eq:unpolarized_light_b} by substituting $a_S^u$ and $s^u_S$ in Equation~\eqref{eq:consistency_unpolarized}.
We then write an equation 
by substituting $\overline{s}_S^u$ with Equation~\eqref{eq:offset_unpolarized} in the combined equation:
\begin{align}\label{eq:final}
k_0  s^\bot = &\left\|  {a^\bot {\rm{exp}({i \varphi^\bot}) }} - {\overline a_S^u {\rm{exp}({i \overline \varphi_S^u}) }} \right\| + \nonumber \\
  & \frac{\overline{a}_S^u{\int_{{\varphi _0}}^\infty a^u_S(\varphi) d\varphi } }{{\left\| {\int_{{\varphi _0}}^\infty  a_S^u (\varphi) \rm{exp}({i \varphi}) } d\varphi \right\|}}.
\end{align}

\mparagraph{Amplitude estimation}
All other variables in Equation~\eqref{eq:final} are already known: $a^\bot$, $\varphi^\bot$, $s^\bot$ from the cross-polarization measurements, $\overline \varphi_S^u$ from Section~\ref{sec:phase-estimation}, and $k_0$ from calibration.
Hence, Equation~\eqref{eq:final} can be reformulated to find amplitude~$\overline a_S^u$ in a closed-form solution. %
We refer to the supplemental document for its analytic solution. %

\subsection{Scattering Removal}
\label{sec:depth-estimate}
\noindent 
Now that we have obtained both phase $\overline \varphi_S^u$ and amplitude $\overline a_S^u$ of unpolarized backscattered light, we are ready to estimate the target depth by computational removing the distortion from backscattered light. %
We simply subtract the phase-amplitude distortion $ {\overline a_S^u {\rm{exp}({i \overline \varphi_S^u}) }}$ from the cross-polarization measurements $ {a^\bot {\rm{exp}({i \varphi^\bot})}}$, resulting in the estimate of unpolarized target light:
\begin{equation}\label{eq:target}
 {a_T^u {\rm{exp}({i \varphi_T^u})}} = {a^\bot {\rm{exp}({i \varphi^\bot}) }} - {\overline a_S^u {\rm{exp}({i \overline \varphi_S^u})}}.
\end{equation}
We can recover the target phase shift, corresponding to its depth, without backscattered distortion as
\begin{equation}\label{eq:depth}
  \varphi_T^u=\mathrm{angle}\left( {a_T^u {\rm{exp}({i \varphi_T^u})}} \right),
\end{equation}
where $\mathrm{angle}(\cdot)$ is the phase-extraction operator.

\begin{figure}[]
	\centering	
	\includegraphics[width=\linewidth]{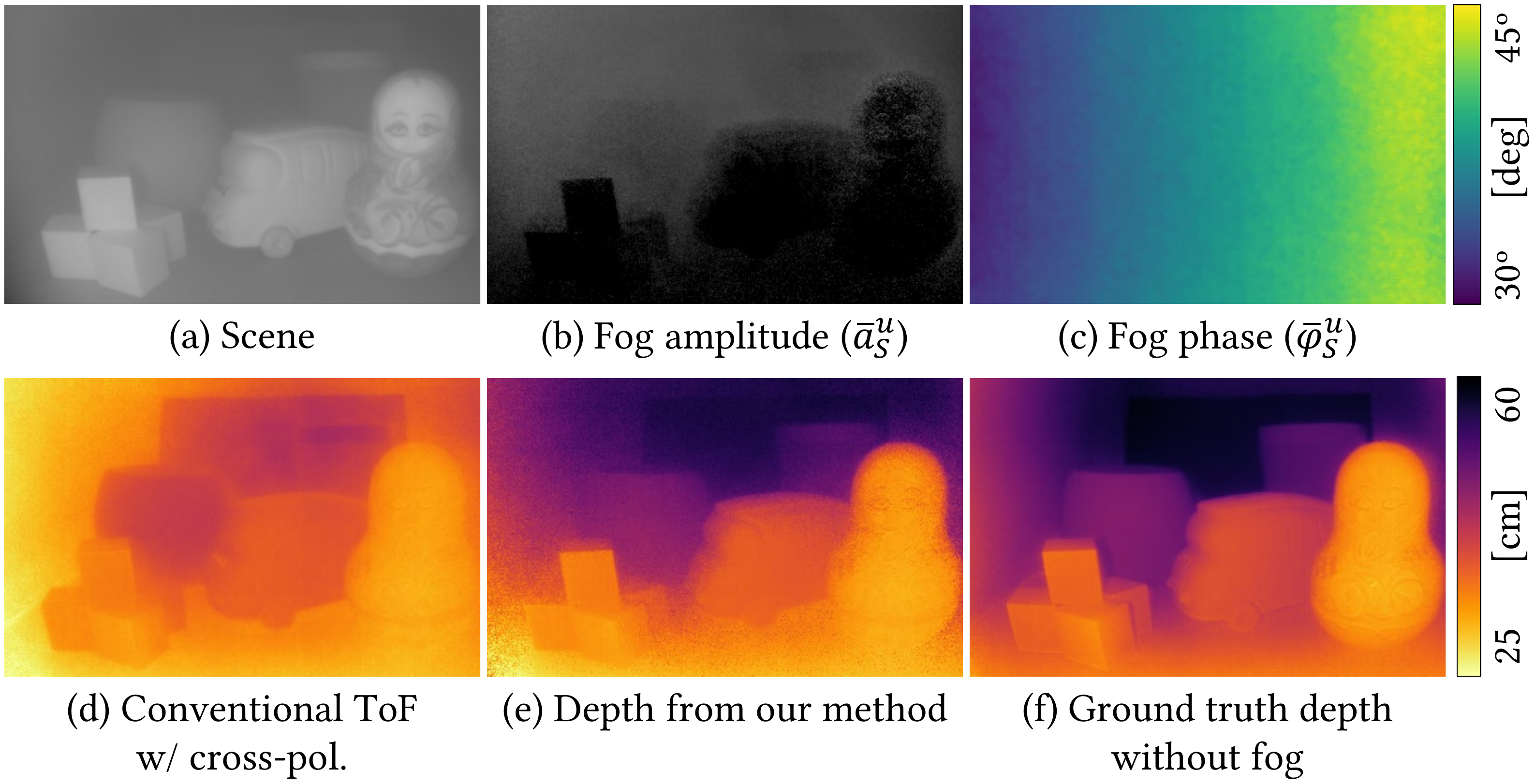}%
	\vspace{-3mm}%
	\caption[]{\label{fig:result_example}%
    We estimate the amplitude and phase caused by unpolarized backscattered light due to fog.
    (a) For a challenging scene with dense fog, we show our estimated fog (b) amplitude $\overline a_S^u$ and (c) phase $\overline \varphi_S^u$.
    (d) While conventional depth estimation from cross-polarization setting fails due to the strong scattering effect, subtracting our estimated backscattered light from the cross-polarization measurements allows us to achieve (e) high-quality depth imaging.
	}%
	\vspace{-6mm}
\end{figure}

\section{Results}
\label{sec:results}

\mparagraph{Experiment details}
Figure~\ref{fig:system_configuration} shows our experimental setup.
We use an indirect ToF camera, Melexis VGA ToF sensor (MLX75027), of which modulation frequency is 80\,MHz and the original spatial resolution of 640\,$\times$\,480.
We use film-based near-infrared (NIR) linear polarizers in front of the ToF illumination and the detector.
The detector-side polarizer is installed on a rotation mount of Thorlabs K10CR1.
Due to the rotation stage's occlusion, we use 380\,$\times$\,240 center crops of the captured images.
To test imaging through scattering media, we install an experimental setup consisting of a 70\,cm\,$\times$\,38\,cm dark chamber, in which we place scene objects.
We generate artificial fog using an off-the-shelf fog generator.
This stable scattering media allows apple-to-apple comparisons between different methods. %
For target scenes, we placed objects of diverse shape and appearance in the chamber and material examples include plastic, acrylic paint, wood, ceramic, or fabric.

\mparagraph{3D imaging through fog}
Figure~\ref{fig:result_example} shows that our polarimetric imaging allows us to see through dense fog and estimate an accurate depth map.
Our method directly estimates the amplitude and phase distortion caused by unpolarized backscattered light.
Figures~\ref{fig:result_example}(b) and (c) show that the estimated fog amplitude and phase match with our capture environment: amplitude is high (brighter in the figure) for far pixels and phase is small for left pixels.
Note that our ToF illumination is located on the left side of the sensor, making the left pixels have smaller phases than the right pixels. %

\begin{figure}[t]
\vspace{-4mm}
	\centering	
	\includegraphics[width=\linewidth]{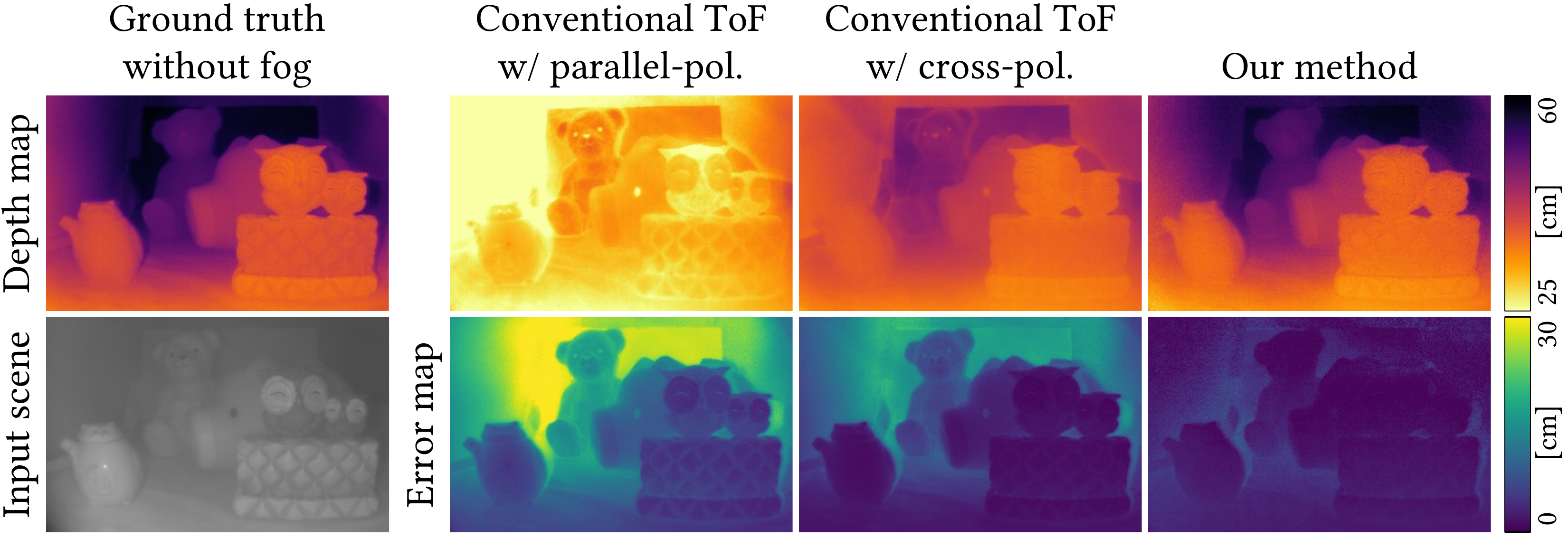}%
	\vspace{-3mm}%
	\caption[]{\label{fig:result_parallel_cross}%
Our method achieves accurate 3D imaging under scattering media compared to conventional ToF imaging with parallel-polarization and cross-polarization configurations.
	}%
	\vspace{-2mm}
\end{figure}

\mparagraph{Depth accuracy}
We evaluate the effectiveness of our method by measuring depth-estimation error.
We first obtain the ground-truth depth without generating any fog as shown in Figure~\ref{fig:result_parallel_cross}.
We capture six scenes each with three fog densities. 
Our quantitative results shown in Table \ref{table:comparison_density} are averaged over the dataset.
While conventional ToF imaging using parallel-polarization and cross-polarization configurations fails to handle the dense scattering environment, our method enables accurate depth reconstruction. 

\begin{table}[t]
	\vspace{-1mm}
	\caption[]{\label{table:comparison_density}%
	Depth accuracy for diverse fog densities. We outperform conventional ToF imaging with cross polarization for all tested environments. %
	}%
	\vspace{-3mm}
	\centering
	\resizebox{0.7\columnwidth}{!}{%
	\begin{tabular}{c|r|r|r}
		\hline
    & \multicolumn{3}{c}{Fog density} \\ \cline{2-4}
	RMSE [cm]   & Thin & Medium & Thick  \\  \hline \hline
	Conventional ToF & 2.78 & 5.46 & 9.11   \\ \hline
	Ours & \textbf{1.71} & \textbf{1.65} & \textbf{2.59}   \\ \hline
	\end{tabular}}
	\vspace{-5mm}
\end{table}

\begin{figure}[t]
	\vspace{-3mm}%
	\centering	
	\includegraphics[width=0.9\linewidth]{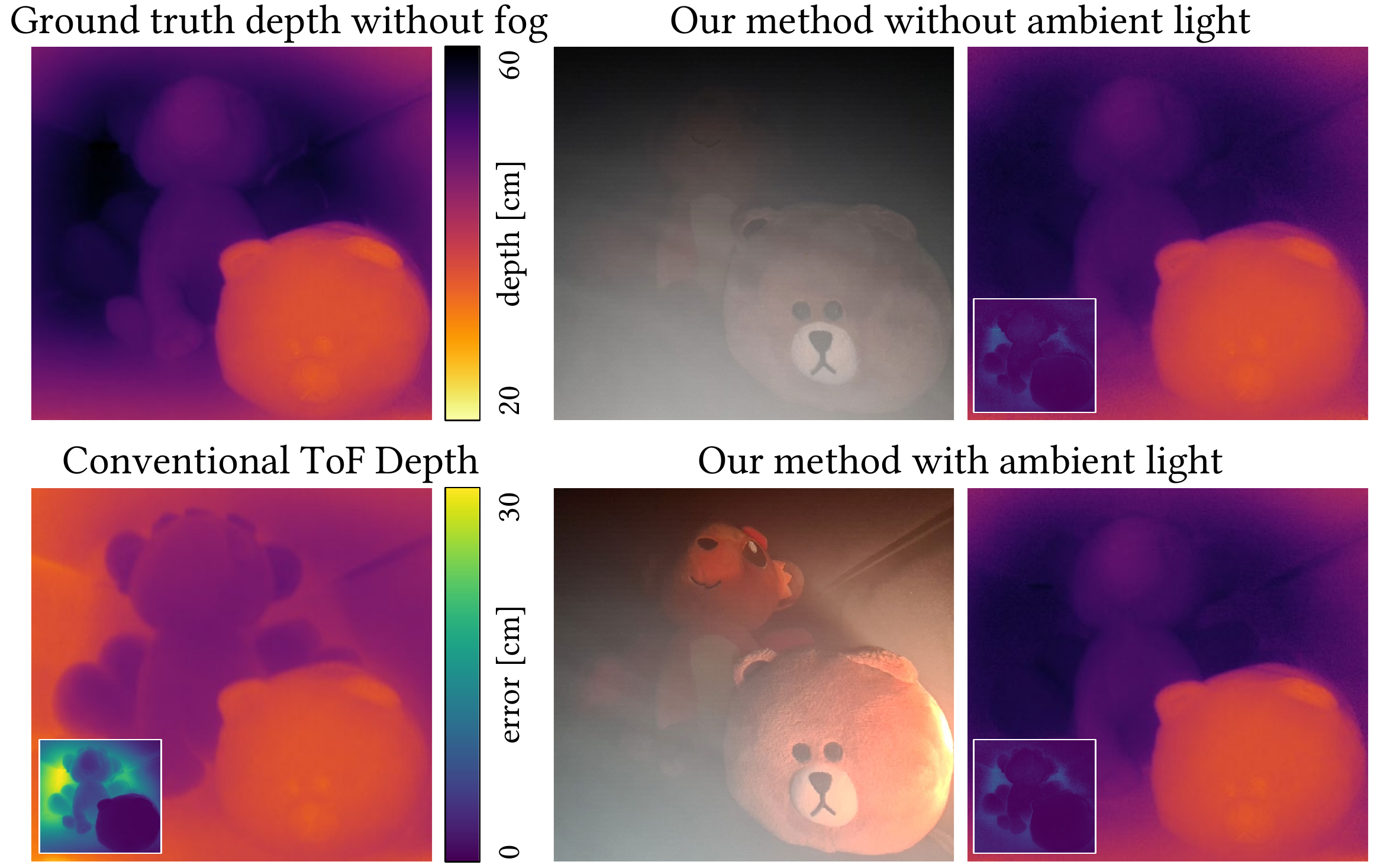}%
	\vspace{-3mm}%
	\caption[]{\label{fig:result_ambient_light}%
	Impact of ambient illumination. Our method can estimate high-accuracy depth information through scattering media as it accounts for ambient illumination when calculating phase. Insets show depth errors.}%
	\vspace{-3mm}
\end{figure}

\mparagraph{Impact of ambient light}
Our method captures an ambient light image without ToF illumination and filters out the ambient contribution to the four-phase correlation measurements like other conventional TOF cameras such as PMD sensors. 
Figure~\ref{fig:result_ambient_light} shows reconstruction results. 
The RMSE of conventional ToF without ambient light is high as 11.62\,cm. 
In contrast, the RMSEs of our results are 1.88\,cm and 2.08\,cm without and with ambient light.

\begin{figure}[t]
	\centering	
	\resizebox{0.75\linewidth}{!}{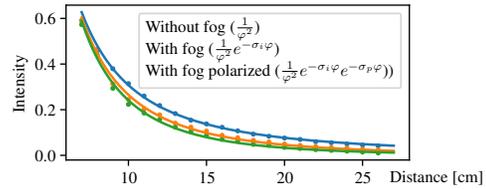}
	\vspace{-3mm}
	\caption[]{\label{fig:evaluation_decay} Measured and fitted intensity decay with respect to distance without and with the fog. We also measure intensity decay with/without a linear polarizer through the fog. Dots represent measurements and lines describe fitting with our decay model.
	\vspace{-3mm}
}
\end{figure}

\mparagraph{Decaying factor}
We validate the amplitude decay model Equation~\eqref{eq:amplitude_of_scattering_polarized} by capturing the energy of light through different distances, as shown in Figure~\ref{fig:evaluation_decay}. 
Intensity decay without fog follows the inverse-square law of emitted light. 
As our system has a linear polarizer in front of the light source, 
we measure intensity with/without a linear polarizer parallel to the sensor's polarizer in front of the camera. 
Our exponential decay model correctly represents the measured data.

\begin{figure}[pt]
	\centering	
	\includegraphics[width=0.95\linewidth]{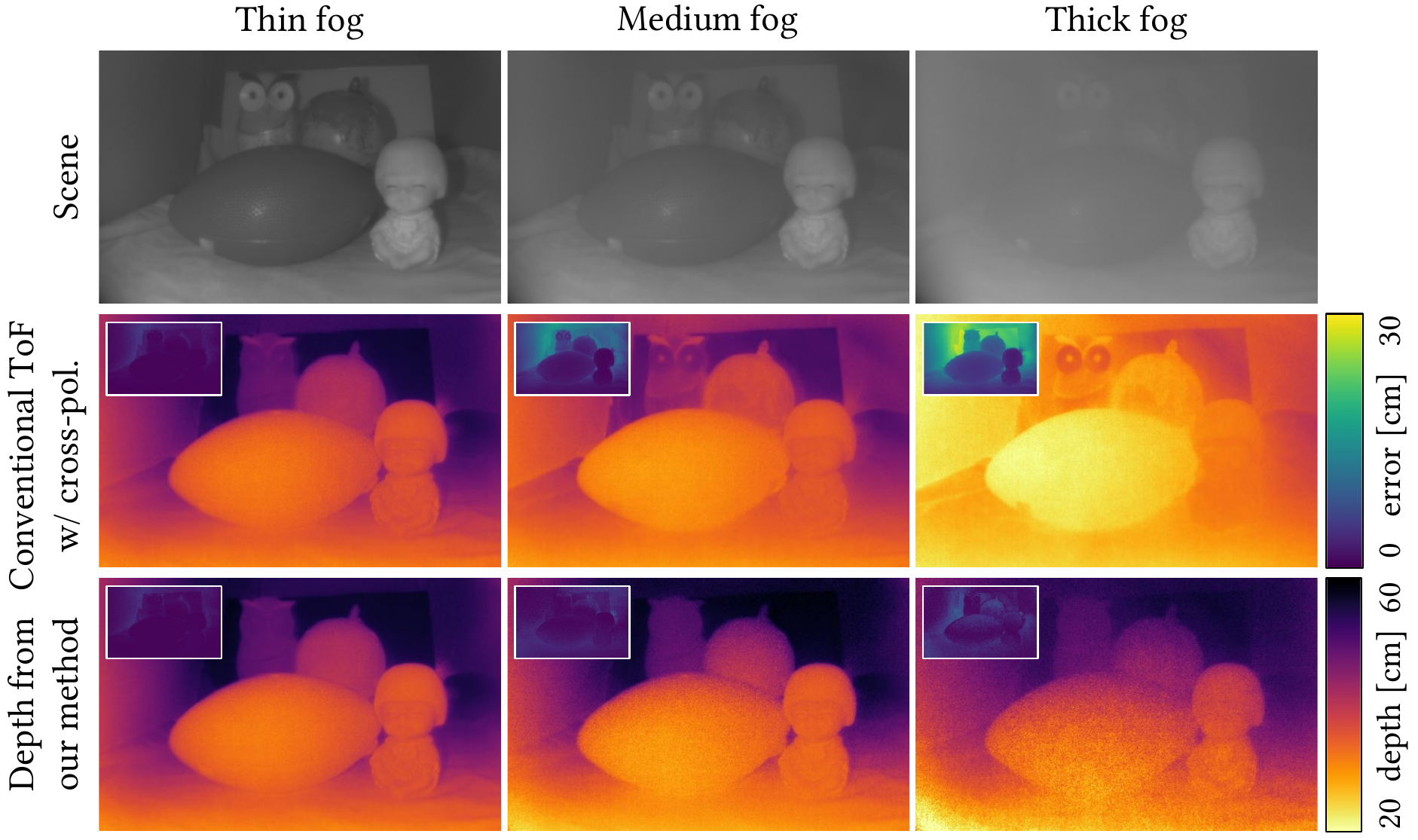}%
	\vspace{-3mm}%
	\caption[]{\label{fig:result_phase_density}%
	Our method reconstructs accurate depth for thin, medium, and thick fog environments. Conventional ToF imaging (with cross-polarization input) suffers from limited scene visibility. %
The estimated amplitude of backscattered unpolarized light becomes higher for denser fog aligned with the environment. %
	}%
	\vspace{-6mm}
\end{figure}

\begin{figure*}[hpt]
	\vspace{-3mm}
	\centering	
	\includegraphics[width=0.88\linewidth]{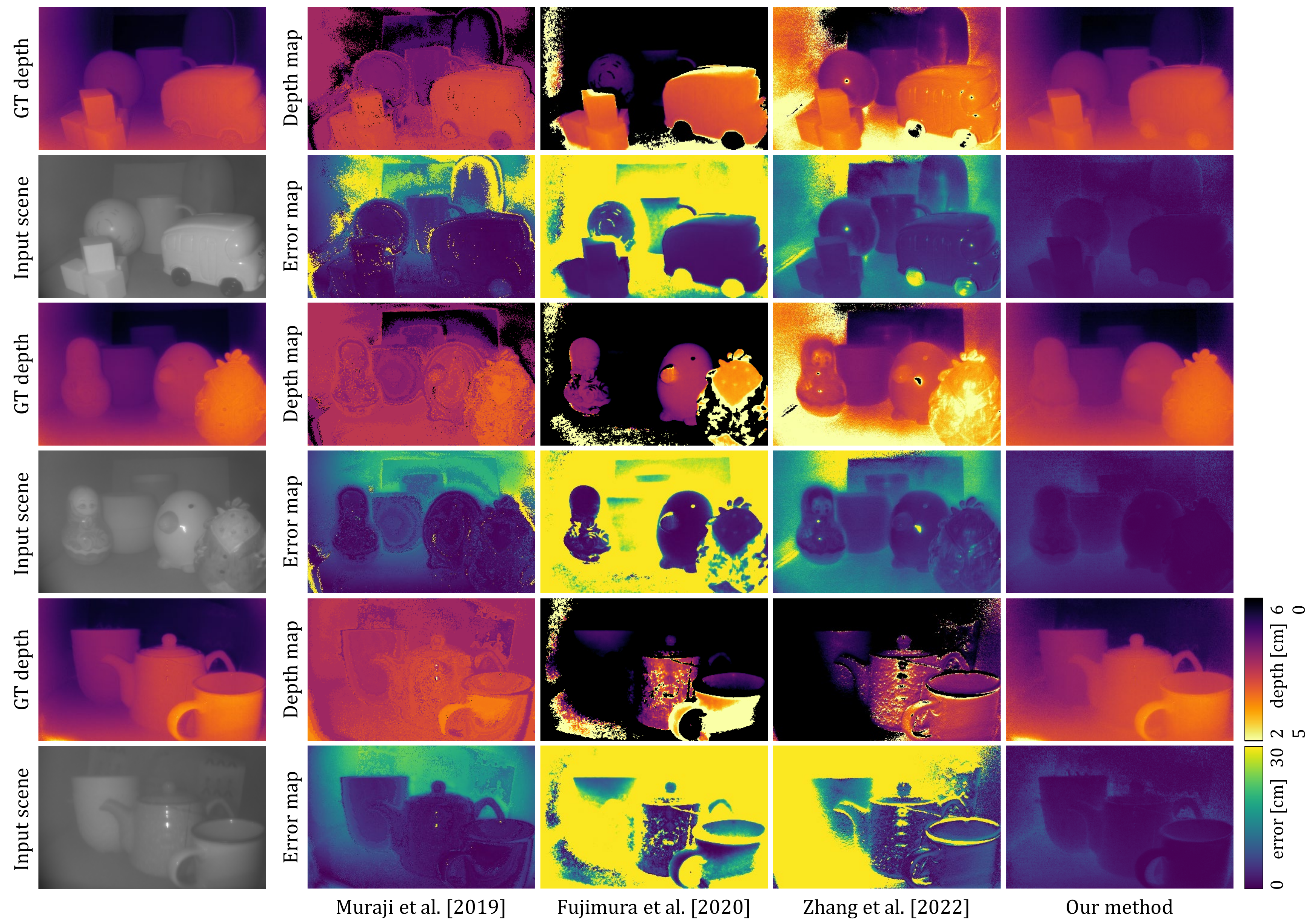}%
	\vspace{-3mm}%
	\caption[]{\label{fig:result_comparison}%
	We compare our method to state-of-the-art scattering-aware ToF methods.  Our method outperforms the previous approaches by a large margin thanks to our polarimetric-ToF image formation.
	}%
	\vspace{-6mm}
\end{figure*}

\mparagraph{Robustness against fog density}
We analyze the robustness of our method against fog density by capturing depth maps with different fog densities.
We adjust the fog generator to realize three different fog densities: thin, medium, and thick.
See Figure \ref{fig:result_phase_density} for qualitative results and Table~\ref{table:comparison_density} for quantitative results.
Our method achieves accurate depth imaging across diverse fog densities, whereas conventional cross-polarization ToF imaging degrades its performance for the medium fog and completely fails in dense fog.

\mparagraph{Comparison}
We compare our method with state-of-the-art ToF imaging methods designed for foggy environments~\cite{muraji2019depth, fujimura2020simultaneous, zhang2022time}.
We capture ToF measurements at parallel- and cross-polarization configurations with multiple ToF modulation frequencies of 40MHz, 50MHz, 60MHz, 70MHz, and 80MHz, in order to provide inputs for the compared methods.
We also compute the depth errors for all the tested scenes and methods by capturing ground-truth scenes without fog.
Figure~\ref{fig:result_comparison} shows the estimated depth maps and their corresponding depth errors are shown in Table~\ref{table:comparison_result}.
Fujimura et al.~\cite{fujimura2020simultaneous} and Zhang et al.~\cite{zhang2022time} requires a background region without any object in the captured, resulting in missing depth estimates for the background pixels.
Moreover, while Zhang et al.~\cite{zhang2022time} assumes that the DOP phasor is uniformly constant over the scene and reflected light is unpolarized, which often does not hold in practice~\cite{baek2018simultaneous}.
In contrast, our method unlocks the previous assumptions, resulting in accurate depth estimation under the scattering medium. 
Muraji et al.~\cite{muraji2019depth} uses multiple ToF modulation frequencies to reduce the impact of scattering media on scene visibility.
Unfortunately, they assume that scene geometry is flat, which does not hold in practice including our tested scenes. %
Our method outperforms all the baseline methods both qualitatively and quantitatively.
Compared to a structured-light-based method~\cite{naik2015light}, our proposed method had benefits in that structured light suffers from large scattering volume such as fog and should capture more than dozens of images, e.g., 25 captures in \cite{naik2015light}.

\begin{table}[pt]
	\caption[]{\label{table:comparison_result}%
We evaluate the depth-estimation accuracy by computing the relative depth error~\cite{geiger2013vision} and the RMSE values.
	Our method outperforms the state-of-the-art ToF imaging methods~\cite{muraji2019depth,fujimura2020simultaneous,zhang2022time}.
	}%
	\vspace{-3mm}
	\resizebox{\columnwidth}{!}{%
	\begin{tabular}{c|r|r|r|r}
		\hline
	Method   & Muraji et al. & Fujimura et al. & Zhang et al. & Ours \\  \hline \hline
	Rel. error & 0.087 & 0.658 & 0.140  & \textbf{0.021} \\
	Std. dev. & 5.10 & 31.92 & 7.52  & \textbf{1.10} \\ 
	RMSE [cm] & 5.73 & 18.64 & 8.17  & \textbf{1.31} \\ \hline
	\end{tabular}}
	\vspace{-7mm}
\end{table}

\section{Conclusion}
\label{sec:conclusion}
\noindent
We have presented a polarimetric iToF imaging method that enables robust depth estimation even in the presence of scattering media. Our approach relies on a novel computational model that incorporates scattering-aware polarimetric phase measurements. Through experimentation, we showcase accurate 3D imaging in dense fog using a polarimetric iToF camera, surpassing the capabilities of other iToF methods. In the future, we aim to leverage micro-polarizer pixel arrays on iToF imaging for dynamic scene capture and to devise an efficient denoising algorithm for the reduced photons.

\appendix
\section*{Acknowledgements}
\noindent Min H.~Kim acknowledges the MSIT/IITP of Korea (RS-2022-00155620, 2022-0-00058, and 2017-0-00072), SK Hynix, and the Samsung Research Funding Center (SRFC-IT2001-04) for developing partial 3D imaging algorithms, in addition to the support of the NIRCH of Korea (2021A02P02-001), Samsung Electronics, and Microsoft Research Asia.
Seung-Hwan Baek acknowledges the support from Samsung Research Funding Center (SRFC-IT1801-52), Samsung Electronics, and Korea NRF grants (2022R1A6A1A03052954, RS-2023-00211658).

\clearpage

{\small
\bibliographystyle{ieee_fullname}
\bibliography{bibliography}

\begin{thebibliography}{10}\itemsep=-1pt

\bibitem{agresti2018deep}
Gianluca Agresti and Pietro Zanuttigh.
\newblock Deep learning for multi-path error removal in tof sensors.
\newblock In {\em Proceedings of the European Conference on Computer Vision
  (ECCV)}, 2018.

\bibitem{baek2021polarimetric}
Seung-Hwan Baek and Felix Heide.
\newblock Polarimetric spatio-temporal light transport probing.
\newblock {\em ACM Transactions on Graphics (TOG)}, 40(6):1--18, 2021.

\bibitem{baek2022all}
Seung-Hwan Baek and Felix Heide.
\newblock All-photon polarimetric time-of-flight imaging.
\newblock In {\em Proceedings of the IEEE/CVF Conference on Computer Vision and
  Pattern Recognition}, pages 17876--17885, 2022.

\bibitem{baek2018simultaneous}
Seung-Hwan Baek, Daniel~S Jeon, Xin Tong, and Min~H Kim.
\newblock Simultaneous acquisition of polarimetric svbrdf and normals.
\newblock {\em ACM Trans. Graph.}, 37(6):268--1, 2018.

\bibitem{baek2020image}
Seung-Hwan Baek, Tizian Zeltner, Hyunjin Ku, Inseung Hwang, Xin Tong, Wenzel
  Jakob, and Min~H Kim.
\newblock Image-based acquisition and modeling of polarimetric reflectance.
\newblock {\em ACM Trans. Graph.}, 39(4):139, 2020.

\bibitem{chenault2000polarization}
David~B Chenault and J~Larry Pezzaniti.
\newblock Polarization imaging through scattering media.
\newblock In {\em Polarization Analysis, Measurement, and Remote Sensing III},
  volume 4133, pages 124--133. International Society for Optics and Photonics,
  2000.

\bibitem{fade2014long}
Julien Fade, Swapnesh Panigrahi, Anthony Carr{\'e}, Ludovic Frein, Cyril Hamel,
  Fabien Bretenaker, Hema Ramachandran, and Mehdi Alouini.
\newblock Long-range polarimetric imaging through fog.
\newblock {\em Applied optics}, 53(18):3854--3865, 2014.

\bibitem{freedman2014sra}
Daniel Freedman, Yoni Smolin, Eyal Krupka, Ido Leichter, and Mirko Schmidt.
\newblock Sra: Fast removal of general multipath for tof sensors.
\newblock In {\em European Conference on Computer Vision}, pages 234--249.
  Springer, 2014.

\bibitem{fuchs2010multipath}
Stefan Fuchs.
\newblock Multipath interference compensation in time-of-flight camera images.
\newblock In {\em 2010 20th International Conference on Pattern Recognition},
  pages 3583--3586. IEEE, 2010.

\bibitem{fujimura2020simultaneous}
Yuki Fujimura, Motoharu Sonogashira, and Masaaki Iiyama.
\newblock Simultaneous estimation of object region and depth in participating
  media using a tof camera.
\newblock {\em IEICE Transactions on Information and Systems}, 103(3):660--673,
  2020.

\bibitem{fukao2021polarimetric}
Yoshiki Fukao, Ryo Kawahara, Shohei Nobuhara, and Ko Nishino.
\newblock Polarimetric normal stereo.
\newblock In {\em Proceedings of the IEEE/CVF Conference on Computer Vision and
  Pattern Recognition}, pages 682--690, 2021.

\bibitem{geiger2013vision}
Andreas Geiger, Philip Lenz, Christoph Stiller, and Raquel Urtasun.
\newblock Vision meets robotics: The kitti dataset.
\newblock {\em The International Journal of Robotics Research},
  32(11):1231--1237, 2013.

\bibitem{gilbert1967improvement}
Gary~D Gilbert and John~C Pernicka.
\newblock Improvement of underwater visibility by reduction of backscatter with
  a circular polarization technique.
\newblock {\em Applied Optics}, 6(4):741--746, 1967.

\bibitem{guo2018tackling}
Qi Guo, Iuri Frosio, Orazio Gallo, Todd Zickler, and Jan Kautz.
\newblock Tackling 3d tof artifacts through learning and the flat dataset.
\newblock In {\em Proceedings of the European Conference on Computer Vision
  (ECCV)}, pages 368--383, 2018.

\bibitem{gupta2015phasor}
Mohit Gupta, Shree~K Nayar, Matthias~B Hullin, and Jaime Martin.
\newblock Phasor imaging: A generalization of correlation-based time-of-flight
  imaging.
\newblock {\em ACM Transactions on Graphics (ToG)}, 34(5):1--18, 2015.

\bibitem{jarry1998coherence}
Gilbert Jarry, Elisa Steimer, Vivien Damaschini, Michael Epifanie, Marc
  Jurczak, and Robin Kaiser.
\newblock Coherence and polarization of light propagating through scattering
  media and biological tissues.
\newblock {\em Applied optics}, 37(31):7357--7367, 1998.

\bibitem{jimenez2014modeling}
David Jim{\'e}nez, Daniel Pizarro, Manuel Mazo, and Sira Palazuelos.
\newblock Modeling and correction of multipath interference in time of flight
  cameras.
\newblock {\em Image and Vision Computing}, 32(1):1--13, 2014.

\bibitem{kerker2013scattering}
Milton Kerker.
\newblock {\em The scattering of light and other electromagnetic radiation:
  physical chemistry: a series of monographs}, volume~16.
\newblock Academic press, 2013.

\bibitem{kijima2021time}
Daiki Kijima, Takahiro Kushida, Hiromu Kitajima, Kenichiro Tanaka, Hiroyuki
  Kubo, Takuya Funatomi, and Yasuhiro Mukaigawa.
\newblock Time-of-flight imaging in fog using multiple time-gated exposures.
\newblock {\em Optics Express}, 29(5):6453--6467, 2021.

\bibitem{konnen1985polarized}
GP Konnen and GP K{\"o}nnen.
\newblock {\em Polarized light in nature}.
\newblock CUP Archive, 1985.

\bibitem{lange2001solid}
Robert Lange and Peter Seitz.
\newblock Solid-state time-of-flight range camera.
\newblock {\em IEEE Journal of quantum electronics}, 37(3):390--397, 2001.

\bibitem{lei2020polarized}
Chenyang Lei, Xuhua Huang, Mengdi Zhang, Qiong Yan, Wenxiu Sun, and Qifeng
  Chen.
\newblock Polarized reflection removal with perfect alignment in the wild.
\newblock In {\em Proceedings of the IEEE/CVF Conference on Computer Vision and
  Pattern Recognition}, pages 1750--1758, 2020.

\bibitem{lewis1999backscattering}
Gareth~D Lewis, David~L Jordan, and P~John Roberts.
\newblock Backscattering target detection in a turbid medium by polarization
  discrimination.
\newblock {\em Applied Optics}, 38(18):3937--3944, 1999.

\bibitem{marco2017deeptof}
Julio Marco, Quercus Hernandez, Adolfo Munoz, Yue Dong, Adrian Jarabo, Min~H
  Kim, Xin Tong, and Diego Gutierrez.
\newblock Deeptof: off-the-shelf real-time correction of multipath interference
  in time-of-flight imaging.
\newblock {\em ACM Transactions on Graphics (ToG)}, 36(6):1--12, 2017.

\bibitem{muraji2019depth}
Takeshi Muraji, Kenichiro Tanaka, Takuya Funatomi, and Yasuhiro Mukaigawa.
\newblock Depth from phasor distortions in fog.
\newblock {\em Optics express}, 27(13):18858--18868, 2019.

\bibitem{mure2007optimized}
James Mure-Dubois and Heinz H{\"u}gli.
\newblock Optimized scattering compensation for time-of-flight camera.
\newblock In {\em Two-and Three-Dimensional Methods for Inspection and
  Metrology V}, volume 6762, page 67620H. International Society for Optics and
  Photonics, 2007.

\bibitem{naik2015light}
Nikhil Naik, Achuta Kadambi, Christoph Rhemann, Shahram Izadi, Ramesh Raskar,
  and Sing Bing~Kang.
\newblock A light transport model for mitigating multipath interference in
  time-of-flight sensors.
\newblock In {\em Proceedings of the IEEE Conference on Computer Vision and
  Pattern Recognition}, pages 73--81, 2015.

\bibitem{nan2009linear}
Zeng Nan, Jiang Xiaoyu, Gao Qiang, He Yonghong, and Ma Hui.
\newblock Linear polarization difference imaging and its potential
  applications.
\newblock {\em Applied optics}, 48(35):6734--6739, 2009.

\bibitem{nayar1993removal}
Shree~K Nayar, X-S Fang, and Terrance Boult.
\newblock Removal of specularities using color and polarization.
\newblock In {\em Proceedings of IEEE Conference on Computer Vision and Pattern
  Recognition}, pages 583--590. IEEE, 1993.

\bibitem{patil2020depth}
Swati~S Patil, Pratik~M Bhade, and VS Inamdar.
\newblock Depth recovery in time of flight range sensors via compressed sensing
  algorithm.
\newblock {\em International Journal of Intelligent Robotics and Applications},
  4(2):243--251, 2020.

\bibitem{rowe1995polarization}
MP Rowe, EN Pugh, J~Scott Tyo, and N Engheta.
\newblock Polarization-difference imaging: a biologically inspired technique
  for observation through scattering media.
\newblock {\em Optics letters}, 20(6):608--610, 1995.

\bibitem{sankaran2002comparative}
Vanitha Sankaran, Joseph~T Walsh~Jr, and Duncan~J Maitland.
\newblock Comparative study of polarized light propagation in biologic tissues.
\newblock {\em Journal of biomedical optics}, 7(3):300--306, 2002.

\bibitem{shashar2004transmission}
Nadav Shashar, Shai Sabbah, and Thomas~W Cronin.
\newblock Transmission of linearly polarized light in seawater: implications
  for polarization signaling.
\newblock {\em Journal of Experimental Biology}, 207(20):3619--3628, 2004.

\bibitem{su2018deep}
Shuochen Su, Felix Heide, Gordon Wetzstein, and Wolfgang Heidrich.
\newblock Deep end-to-end time-of-flight imaging.
\newblock In {\em Proceedings of the IEEE Conference on Computer Vision and
  Pattern Recognition}, pages 6383--6392, 2018.

\bibitem{treibitz2008active}
Tali Treibitz and Yoav~Y Schechner.
\newblock Active polarization descattering.
\newblock {\em IEEE transactions on pattern analysis and machine intelligence},
  31(3):385--399, 2008.

\bibitem{van2015detection}
JD Van~der Laan, DA Scrymgeour, SA Kemme, and EL Dereniak.
\newblock Detection range enhancement using circularly polarized light in
  scattering environments for infrared wavelengths.
\newblock {\em Applied optics}, 54(9):2266--2274, 2015.

\bibitem{wang2020analyzing}
Pengfei Wang, Dekui Li, Xinyang Wang, Kai Guo, Yongxuan Sun, Jun Gao, and
  Zhongyi Guo.
\newblock Analyzing polarization transmission characteristics in foggy
  environments based on the indices of polarimetric purity.
\newblock {\em IEEE Access}, 8:227703--227709, 2020.

\bibitem{wilkie2004analytical}
Alexander Wilkie, Claudia Ulbricht, Robert~F Tobler, Georg Zotti, and Werner
  Purgathofer.
\newblock An analytical model for skylight polarisation.
\newblock In {\em Rendering Techniques}, pages 387--398, 2004.

\bibitem{zeng2018visible}
Xiangwei Zeng, Jinkui Chu, Wenda Cao, Weidong Kang, and Ran Zhang.
\newblock Visible--ir transmission enhancement through fog using circularly
  polarized light.
\newblock {\em Applied Optics}, 57(23):6817--6822, 2018.

\bibitem{zhang2022time}
Yixin Zhang, Xia Wang, Yuwei Zhao, and Yujie Fang.
\newblock Time-of-flight imaging in fog using polarization phasor imaging.
\newblock {\em Sensors}, 22(9):3159, 2022.

\bibitem{zhao2022polarization}
Yuwei Zhao, Xia Wang, Yixin Zhang, Yujie Fang, and BingHua Su.
\newblock Polarization-based approach for multipath interference mitigation in
  time-of-flight imaging.
\newblock {\em Applied Optics}, 61(24):7206--7217, 2022.

\end{thebibliography}
}


\clearpage

\onecolumn

\title{\centering\begin{huge}\textbf{Supplemental Document}\end{huge}\\[20mm]}

\noindent
In this Supplemental Document, we provide details of the analytic solution about the amplitude estimation of unpolarized backscattered light.
We also additional discussion, results and comparisons.

\section{Amplitude Estimation}
\label{sec:equation solution}
\noindent
For completeness, we start by rewriting Equation (16) in the main paper here:
\begin{align}\label{eq:final}
\left\|  {a^\bot {\rm{exp}({i \varphi^\bot}) }} - {\overline a_S^u {\rm{exp}({i \overline \varphi_S^u}) }} \right\| - k_0 s^\bot + \frac{\overline{a}_S^u{\int_{{\varphi _0}}^\infty a^u_S(\varphi) d\varphi } }{{\left\| {\int_{{\varphi _0}}^\infty  a_S^u (\varphi) \rm{exp}({i \varphi}) } d\varphi \right\|}} =0.
\end{align}

\noindent We can further expand this using the Euler equation as:
\begin{equation}
    {\left( {{a^ \bot }\cos \left( {{\varphi ^ \bot }} \right) - \overline a_S^u\cos \left( {\overline \varphi _S^u} \right)} \right)^2} + 
    {\left( {{a^ \bot }\sin \left( {{\varphi ^ \bot }} \right) - \overline a_S^u\sin \left( {\overline \varphi _S^u} \right)} \right)^2} - 
    {\left( {{k_0}{s^ \bot } - \overline a_S^u\frac{{{k_0}}}{{\overline k_S^u}}} \right)^2} = 0.
\end{equation}
We then rearrange the formation with respect to the amplitude $\overline a_S^u$ as 
\begin{equation}
\label{eq:quad}
    {\left( {\overline a_S^u} \right)^2}\left( {1 - {{\left( {\frac{{{k_0}}}{{\overline k_S^u}}} \right)}^2}} \right) - 
    2\overline a_S^u\left[ {{a^ \bot }\left( {\cos \left( {{\varphi ^ \bot }} \right)\cos \left( {\overline \varphi _S^u} \right) + \sin \left( {{\varphi ^ \bot }} \right)\sin \left( {\overline \varphi _S^u} \right)} \right) - \frac{{{{\left( {{k_0}} \right)}^2}}}{{\overline k_S^u}}{s^ \bot }} \right] + 
    {\left( {{a^ \bot }} \right)^2} - {\left( {{k_0}{s^ \bot }} \right)^2}.
\end{equation}
Note that Equation~\eqref{eq:quad} is the quadratic equation for the amplitude $\overline a_S^u$.
The solution for the quadratic equation is given as 
\begin{align}
\overline a_S^u &= \frac{{c_2 + \sqrt {{c_2^2} - c_1c_3} }}{c_1}, \mathrm{where} \nonumber \\
    {c_1} &= \left( {1 - {{\left( {\frac{{{k_0}}}{{\overline k_S^u}}} \right)}^2}} \right), \nonumber\\
    {c_2} &= \left[ {{a^ \bot }\left( {\cos \left( {{\varphi ^ \bot }} \right)\cos \left( {\overline \varphi _S^u} \right) + \sin \left( {{\varphi ^ \bot }} \right)\sin \left( {\overline \varphi _S^u} \right)} \right) - \frac{{{{\left( {{k_0}} \right)}^2}}}{{\overline k_S^u}}s^\bot} \right], \nonumber\\
    {c_3} &= {\left( {{a^ \bot }} \right)^2} - {\left( {{k_0}{s^ \bot }} \right)^2}.
\end{align}

\section{Calibration of Scalar $\alpha$}
\noindent 
While the extinction coefficients $\sigma$, $\sigma_i$, and $\sigma_p$ in our model depend on the fog density, the scalar~$\alpha$ is independent of the density. 
Rather, the scalar~$\alpha$ has a dependency in terms of spatial locations, resulting in per-pixel $\alpha$.
To obtain the spatially-varying $\alpha$, we built a closed chamber where a fog generator creates fog. We then wait for the fog to stabilize within the box physically and capture the foggy scene without objects in order to calibrate the scalar $\alpha$.

\section{Discussions}

\subsection{Optically-thin Scattering Media}  
\noindent There are a couple of assumptions that our method is based on.
First, we assume that diffuse polarization from the surface is weak enough to be negligible $T_\phi^ p \approx 0$ through multiple scattering. 
Second, we assume that light transport occurs through optically thin scattering media, i.e., light scattering reflected from the target surfaces is negligible. 

\subsection{Fog Homogeneity Assumption} 
\noindent In the real-world imaging scenario, a participating medium may consist of heterogeneous particles. However, it is too challenging to formulate the entire interaction among heterogeneous participating particles in an analytical form. To keep the fog estimation problem tractable, we assume the homogeneity of the participating medium, focusing on depth estimation. 

\subsection{2$\pi$ Ambiguity} 
\noindent Indirect ToF can estimate with a phase range from 0 to 2$\pi$. A phase over $2\pi$ is wrapped by 2$\pi$, which limits the depth estimation range. 
The depth limit within a single wrapping count is defined as $\frac{c}{2f}$, where $c$ is the speed of light in the operating medium, and $f$ is the modulation frequency.
Hence, the depth limit can be extended by choosing a lower frequency $f$ for light modulation as long as the sensor's dynamic range allows. 
In practice, for active i-ToF imaging, the maximum depth range is usually limited by the power of the light source rather than the wrapping distance, so we assume that the wrapping problem is not critical for our application.
Moreover, the MHz modulation frequencies used in typical iToF cameras present a small number of phase wraps which can be estimated in a robust manner using a variant of the Chinese remainder theorem.

\subsection{Spatial Regularization} 
\noindent Our main scope is to devise a polarized phasor model for the indirect ToF in scattering media and solve the model analytically to estimate the depth from each pixel's phasor observation. This method is directly applicable to each pixel independently.
Our estimated depth may present noise, which could be improved by plugging in an additional process, such as adding a regularization term when computing the extinction coefficient for each pixel or applying smoothing algorithms such as a guided filter. 
Developing an efficient depth denoising algorithm and phase-reconstruction method would be interesting for future work.

\subsection{Assumption of random polarization} 
\noindent As the skylight is partially polarized depending on the angle between the sun and the observer, i.e., near grazing angles, 
polarized scattering occurs in nature. Also, it can be modeled using a deterministic scattering model of polarization.
However, in our experimental setup, the camera and the active illumination are placed next to each other, looking in the same direction. In this coaxial optics setup, such deterministic scattering rarely occurs.
Random polarization tends to occur through retro-reflection. To our knowledge, the proposed model may not be applicable to the deterministic scattering of polarized light.

\section{Additional Results}
\label{sec:supplemental results}
\noindent
Here, we provide additional qualitative results, comparison, and evaluations.

\subsection{Subsurface scattering and sparse $k$-bounce interference}
\noindent
We aim to capture opaque objects under scattering media, which commonly occurs in practical applications such as autonomous driving under fog and underwater navigation. 
We thus design our method with an assumption that the TOF phase continuously decreases within the environment. 
It would be an interesting future work to extend our method to handle sparse $k$-bounce multi-path and diffuse subsurface scattering.
This is revealed in our experiment of capturing a convex V-groove object under fog.
Our method achieves higher depth accuracy compared to the naive ToF imaging from the RMSE depth error of 2.39\,cm to 0.61\,cm, the estimated inner angle of the V-groove is 97.56$^\circ$, which deviates from the ground-truth 90$^\circ$, due to sparse $k$-bounce multipath interference.

\subsection{Comparison}
\begin{figure*}[h]
	\centering	
	\includegraphics[width=\linewidth]{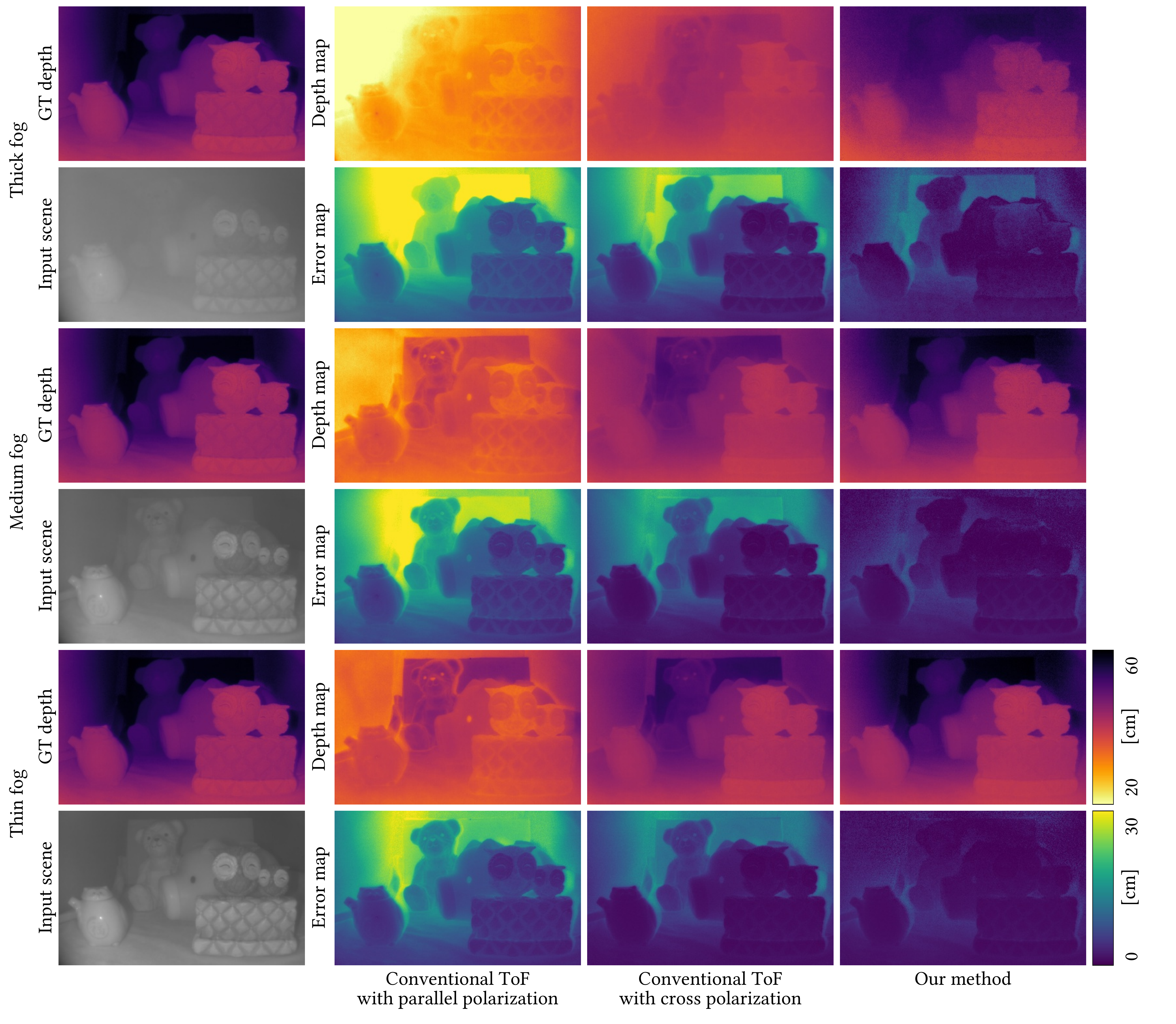}%
	\vspace{-5mm}%
	\caption[]{\label{fig:result11}%
    \emph{Scene: Bear and Owl.} Our method successfully estimates accurate depth for varying degrees of fog density. Note that conventional ToF methods with parallel- and cross-polarization configurations fail to handle the scattering effect.
	}%
\end{figure*}

\begin{figure*}[h]
	\centering	
	\includegraphics[width=\linewidth]{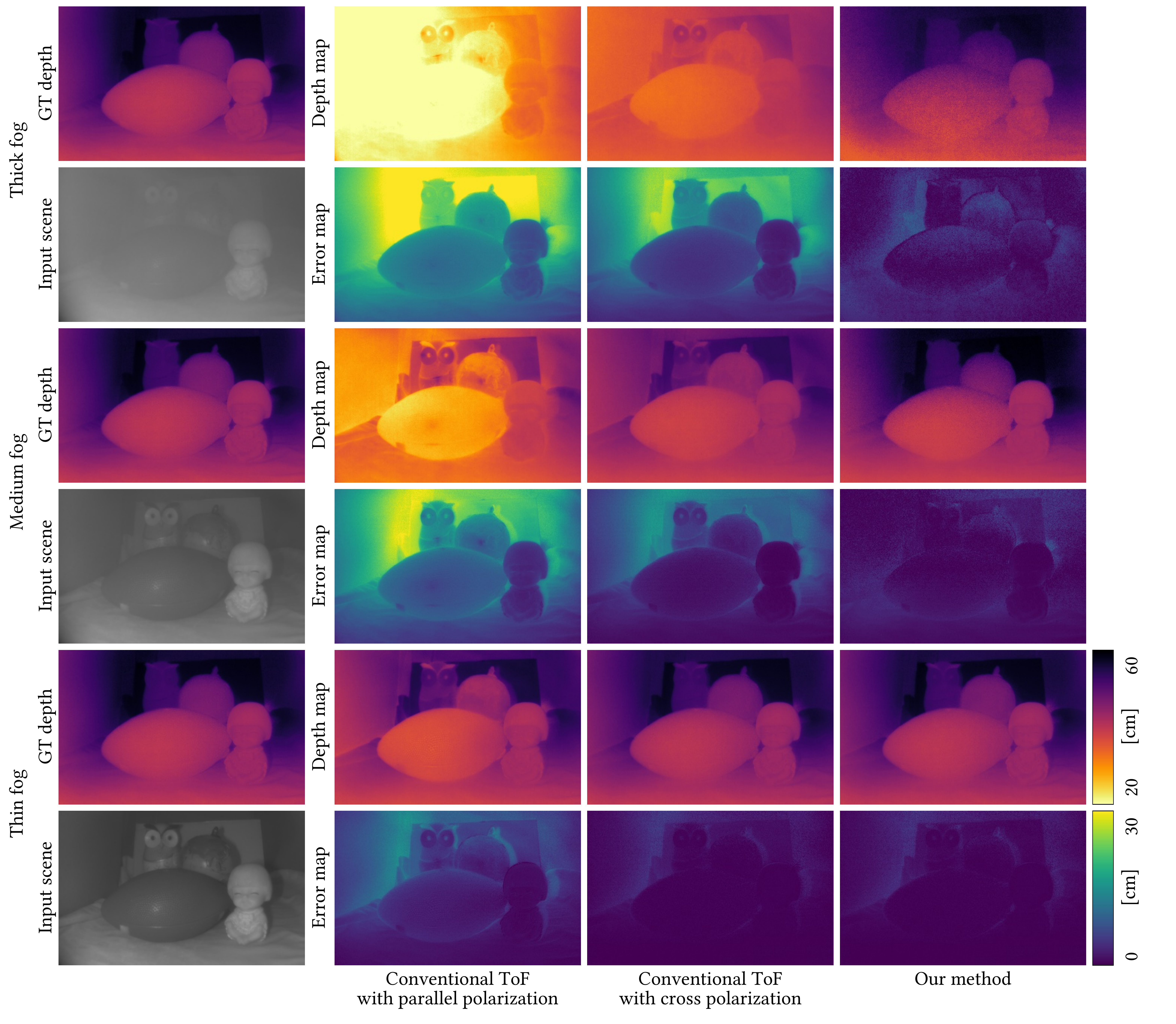}%
	\vspace{-5mm}%
	\caption[]{\label{fig:result12}%
    \emph{Scene: Ball and Earth.} Our method successfully estimates accurate depth for varying degrees of fog density. Note that conventional ToF methods with parallel- and cross-polarization configurations fail to handle the scattering effect.
	}%
\end{figure*}

\begin{figure*}[h]
	\centering	
	\includegraphics[width=\linewidth]{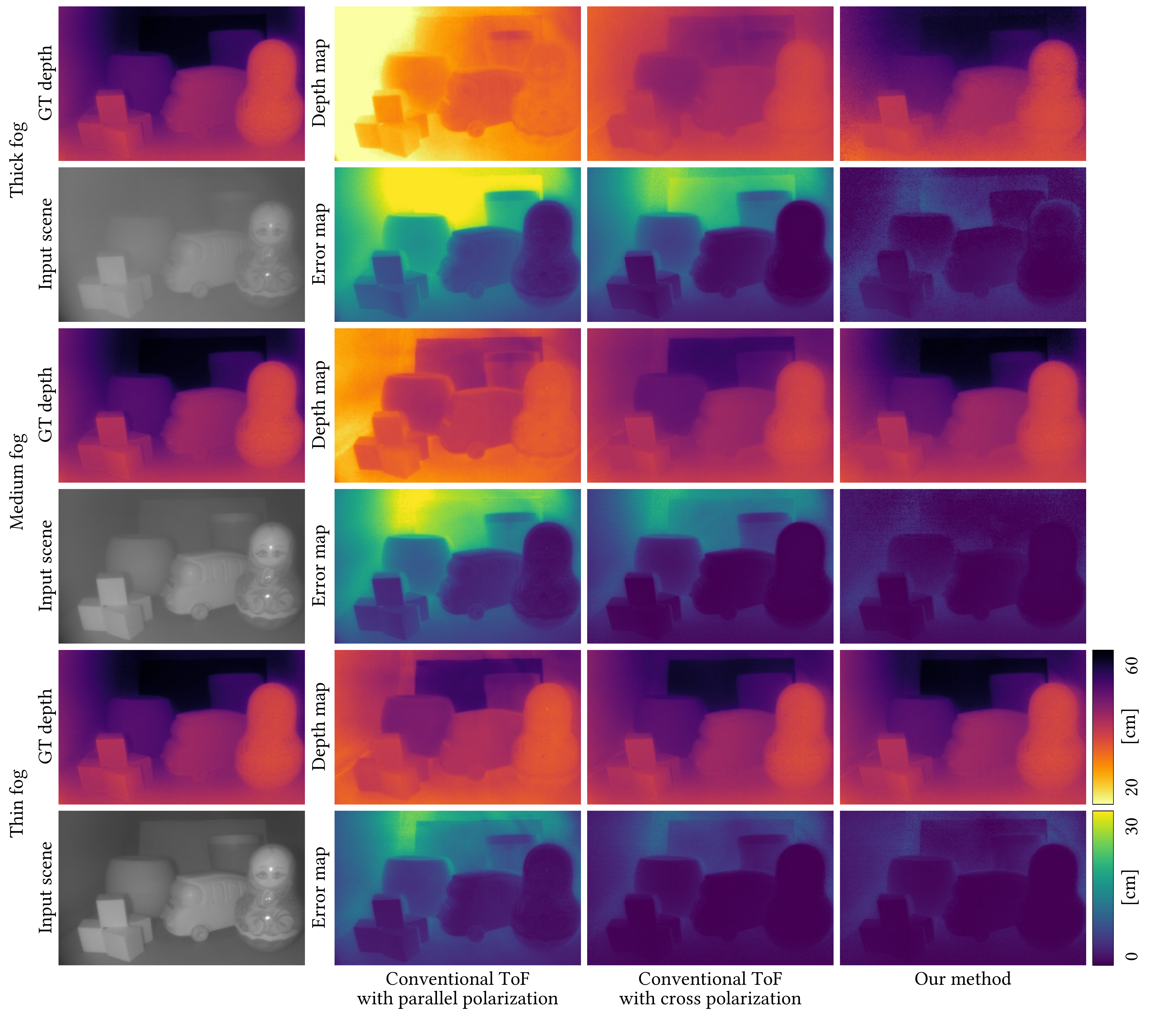}%
	\vspace{-5mm}%
	\caption[]{\label{fig:result13}%
    \emph{Scene: Doll and Block.} Our method successfully estimates accurate depth for varying degrees of fog density. Note that conventional ToF methods with parallel- and cross-polarization configurations fail to handle the scattering effect.
	}%
\end{figure*}

\begin{figure*}[h]
	\centering	
	\includegraphics[width=\linewidth]{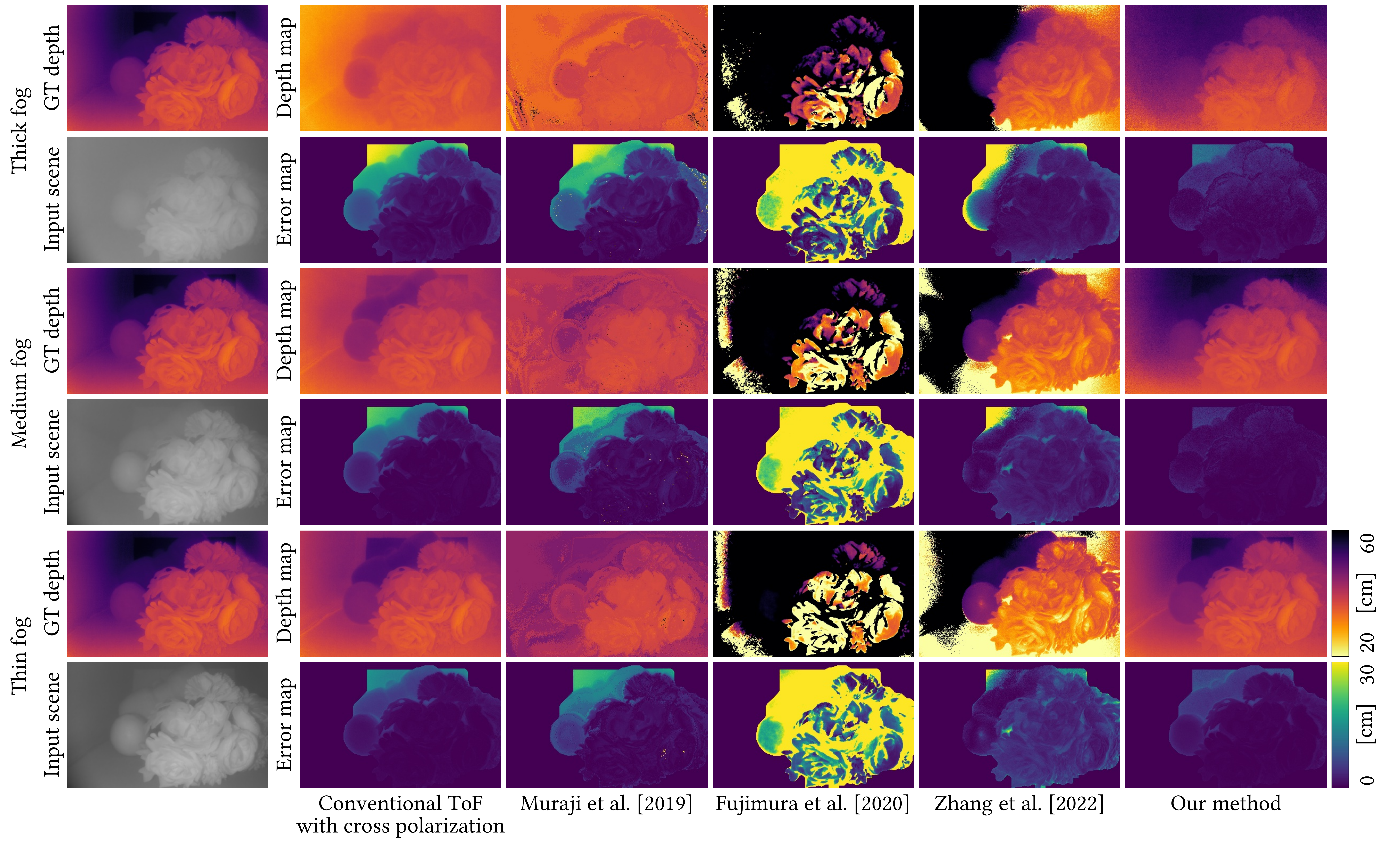}%
	\vspace{-5mm}%
	\caption[]{\label{fig:result21}%
    \emph{Scene: Flower and Orange.} We compare our method to state-of-the-art ToF methods for scattering environments. We outperform all the compared methods.
	}%
\end{figure*}

\begin{figure*}[h]
	\centering	
	\includegraphics[width=\linewidth]{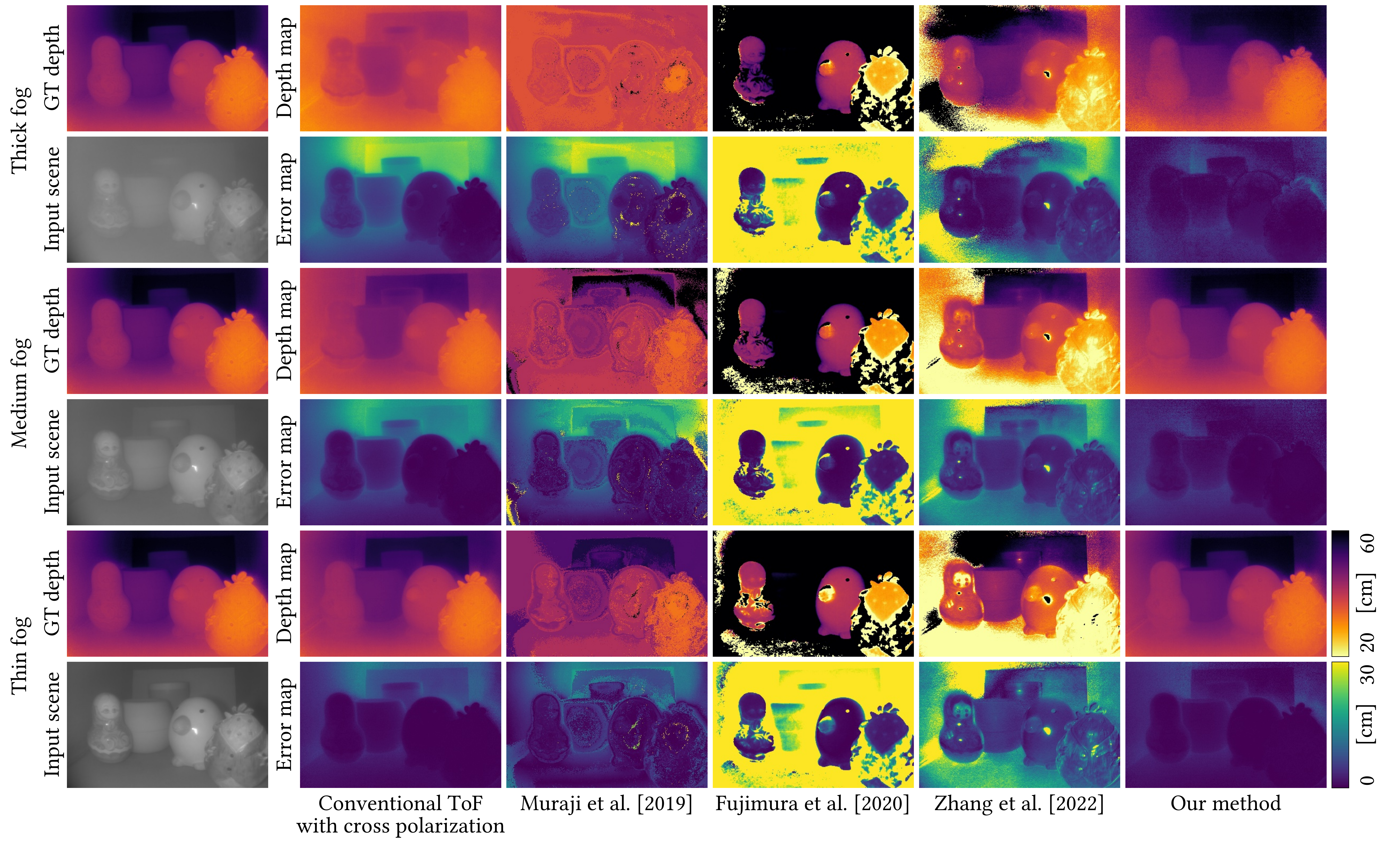}%
	\vspace{-5mm}%
	\caption[]{\label{fig:result22}%
    \emph{Scene: Duck.} We additionally compare our method to state-of-the-art ToF methods for scattering environments.
	}%
\end{figure*}

\begin{figure*}[h]
	\centering	
	\includegraphics[width=\linewidth]{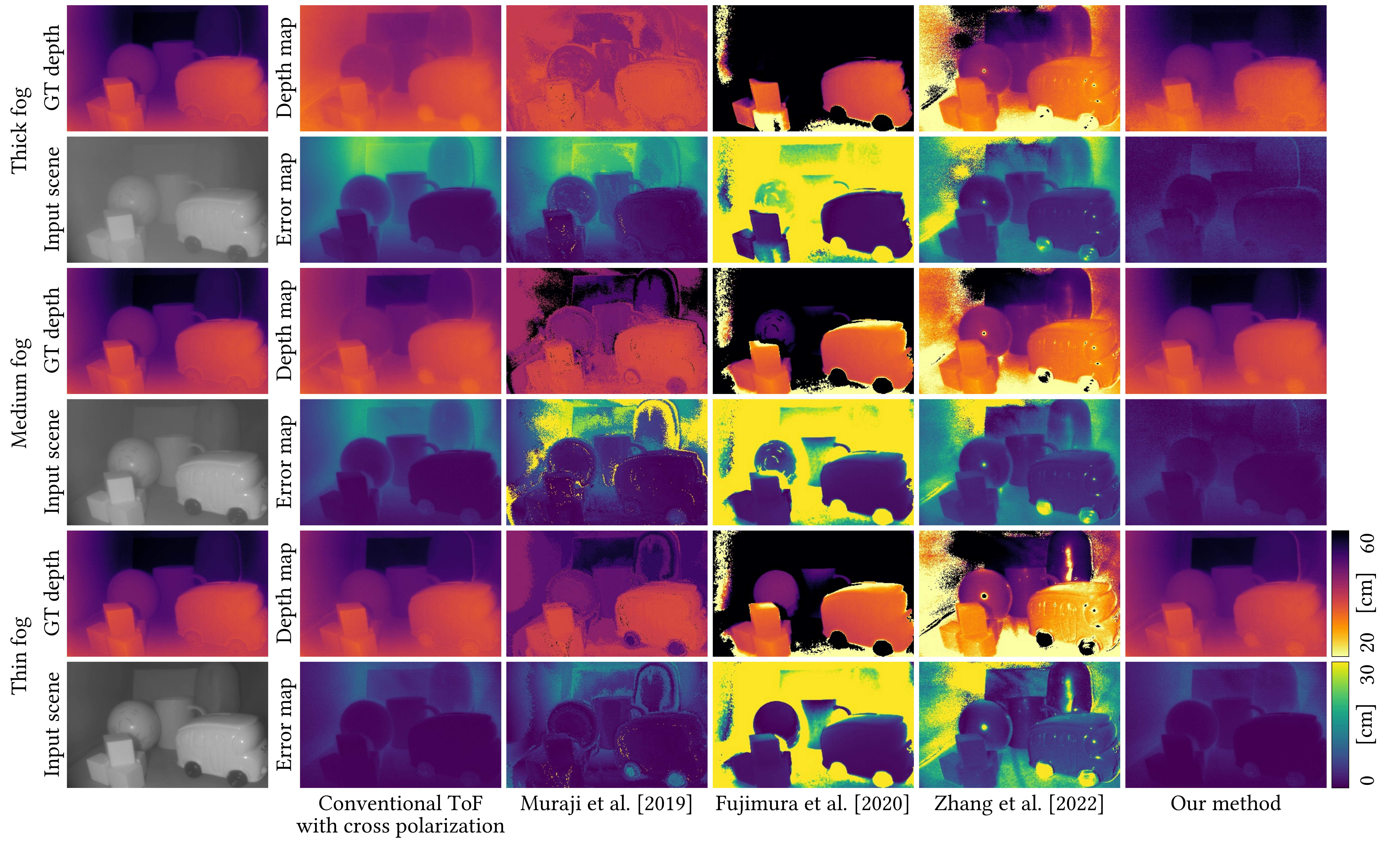}%
	\vspace{-5mm}%
	\caption[]{\label{fig:result23}%
    \emph{Scene: Bus and Earth.} We additionally compare our method to state-of-the-art ToF methods for scattering environments.
	}%
\end{figure*}

\begin{figure*}[h]
	\centering	
	\includegraphics[width=\linewidth]{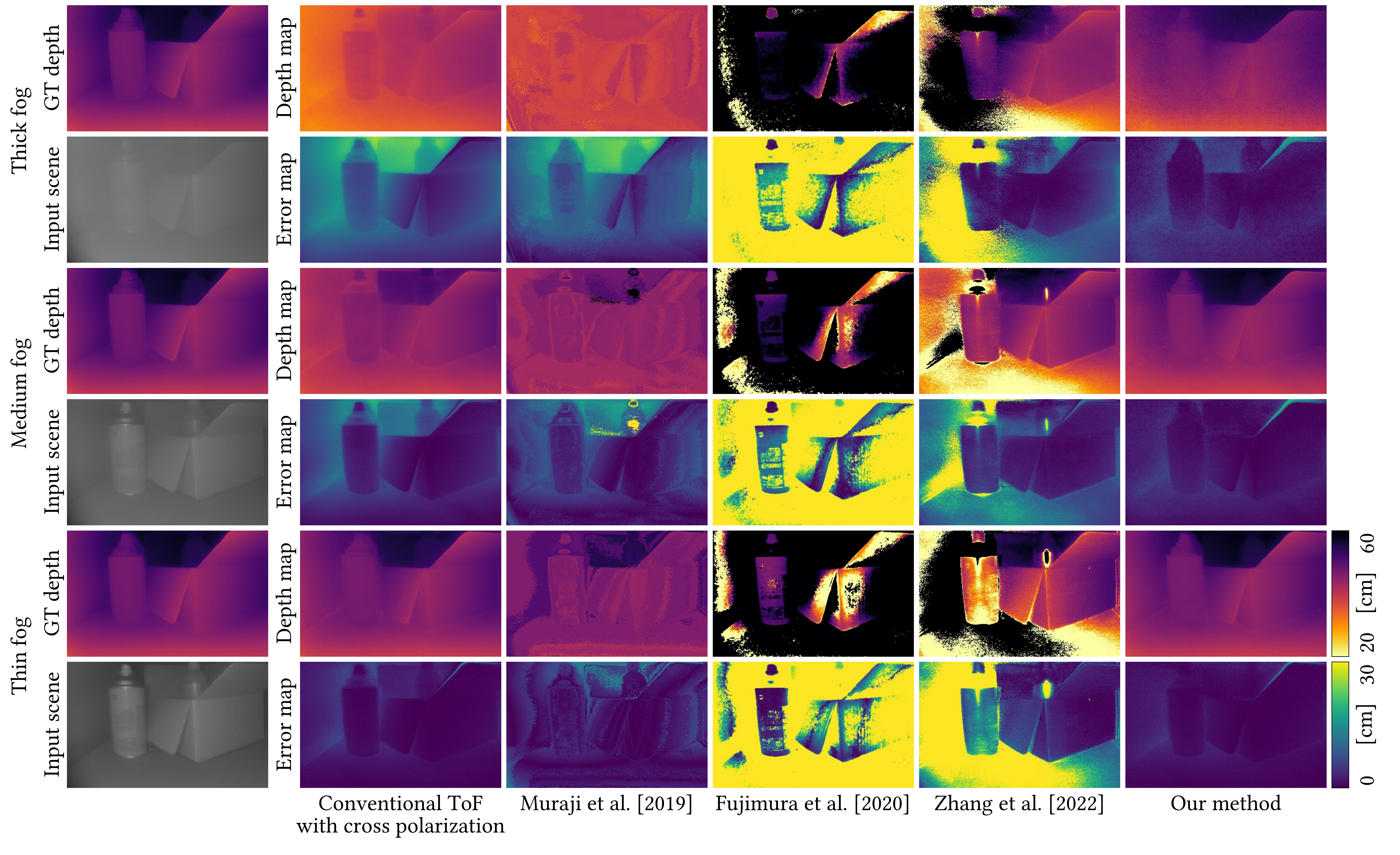}%
	\vspace{-5mm}%
	\caption[]{\label{fig:result24}%
    \emph{Scene: Box and Can.} We additionally compare our method to state-of-the-art ToF methods for scattering environments.
	}%
\end{figure*}

\begin{figure*}[h]
	\centering	
	\includegraphics[width=\linewidth]{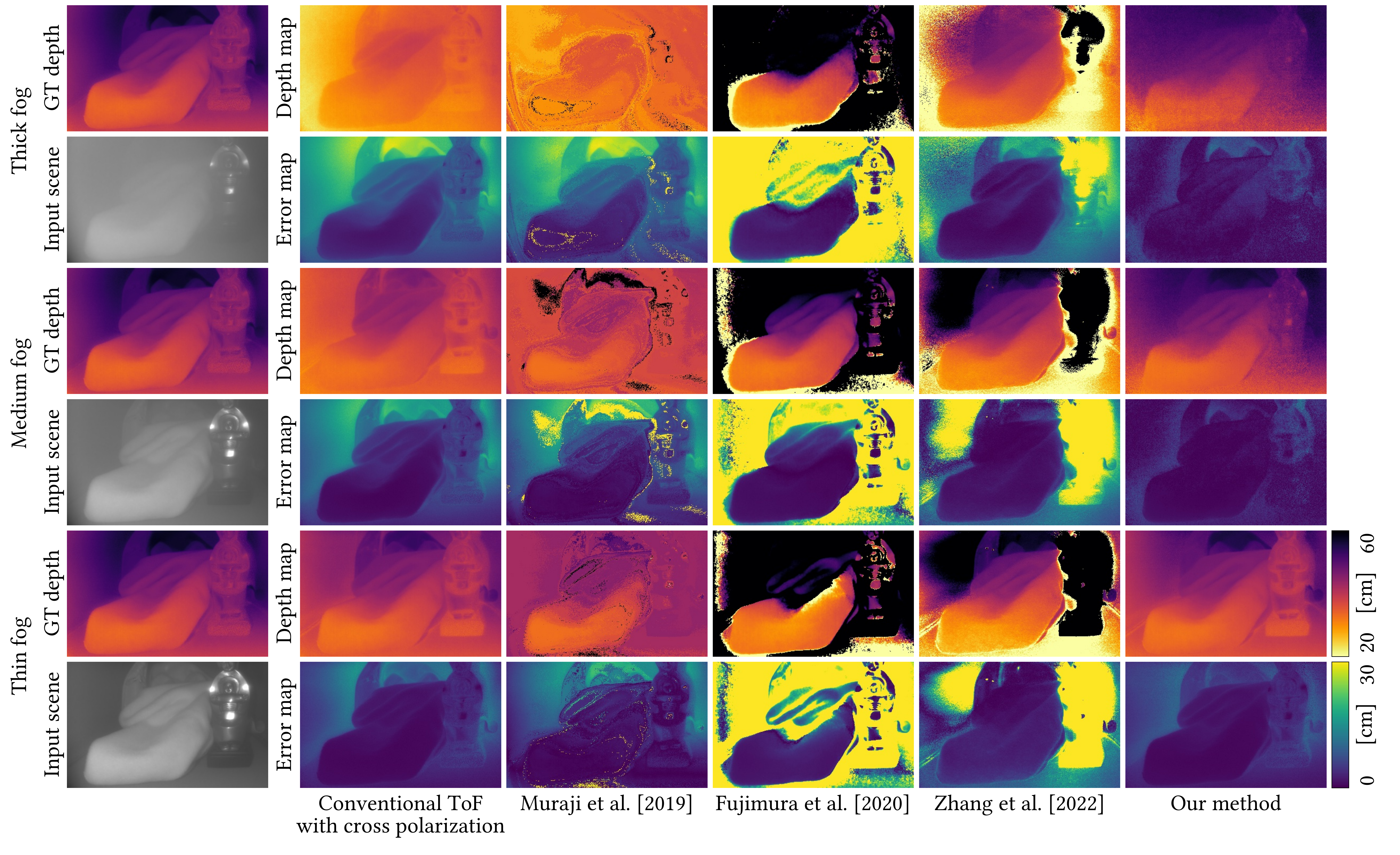}%
	\vspace{-5mm}%
	\caption[]{\label{fig:result25}%
    \emph{Scene: Cloth and Statue.} We additionally compare our method to state-of-the-art ToF methods for scattering environments. 
	}%
\end{figure*}

\noindent
Figures~\ref{fig:result11}, \ref{fig:result12}, and \ref{fig:result13} show our results on varying fog densities compared to parallel- and cross-polarization imaging, where our method outperforms the baselines.
Figures~\ref{fig:result21}, \ref{fig:result22}, \ref{fig:result23}, \ref{fig:result24}, and \ref{fig:result25} demonstrate that our method outperforms the state-of-the-art ToF imaging methods developed for scattering environments.


\begin{figure*}[h]
	\centering	
	\includegraphics[width=\linewidth]{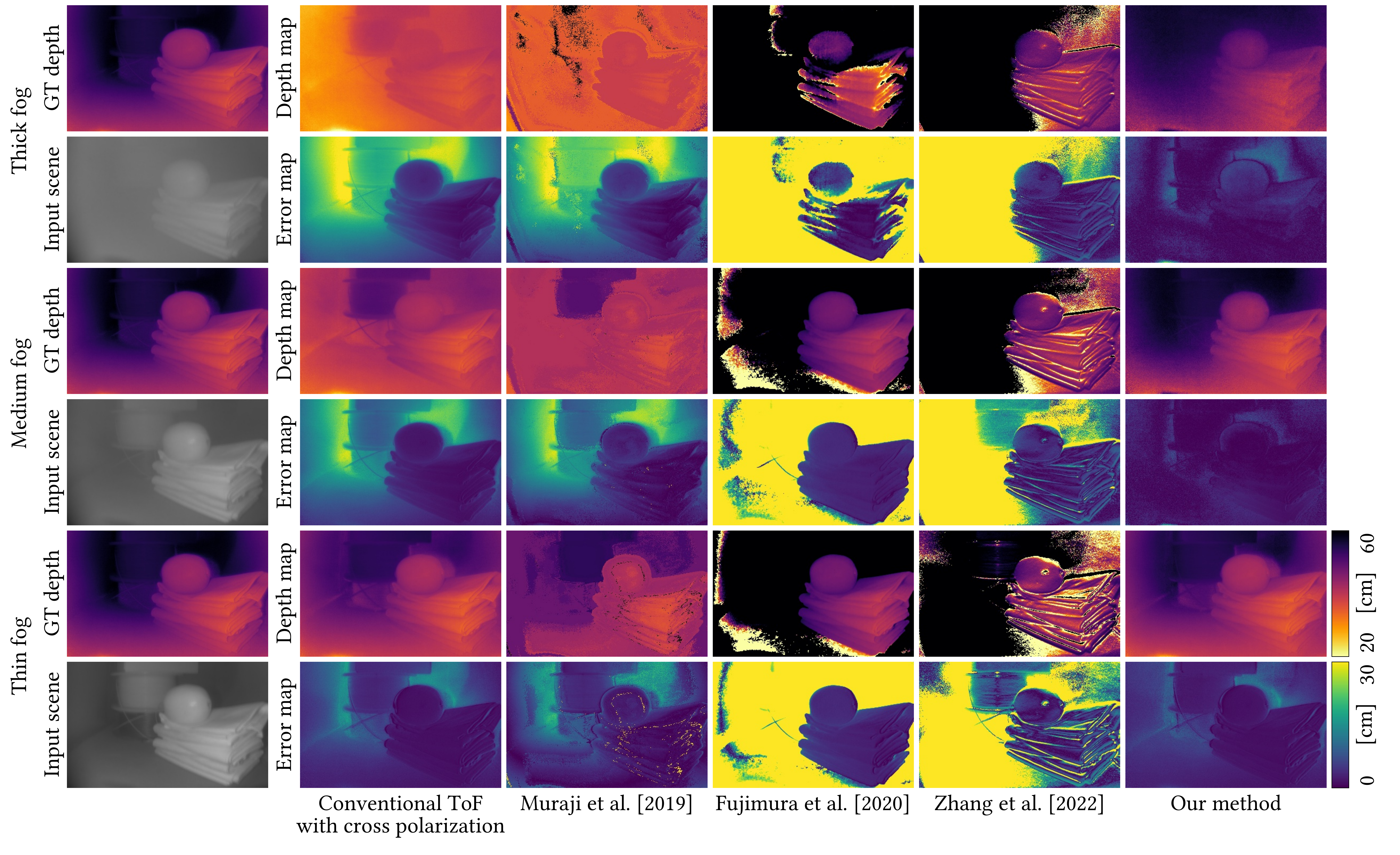}%
	\vspace{-5mm}%
	\caption[]{\label{fig:result26}%
    \emph{Scene: Filament and Cloth.} We additionally compare our method to state-of-the-art ToF methods for scattering environments.
	}%
\end{figure*}

\begin{figure*}[h]
	\centering	
	\includegraphics[width=\linewidth]{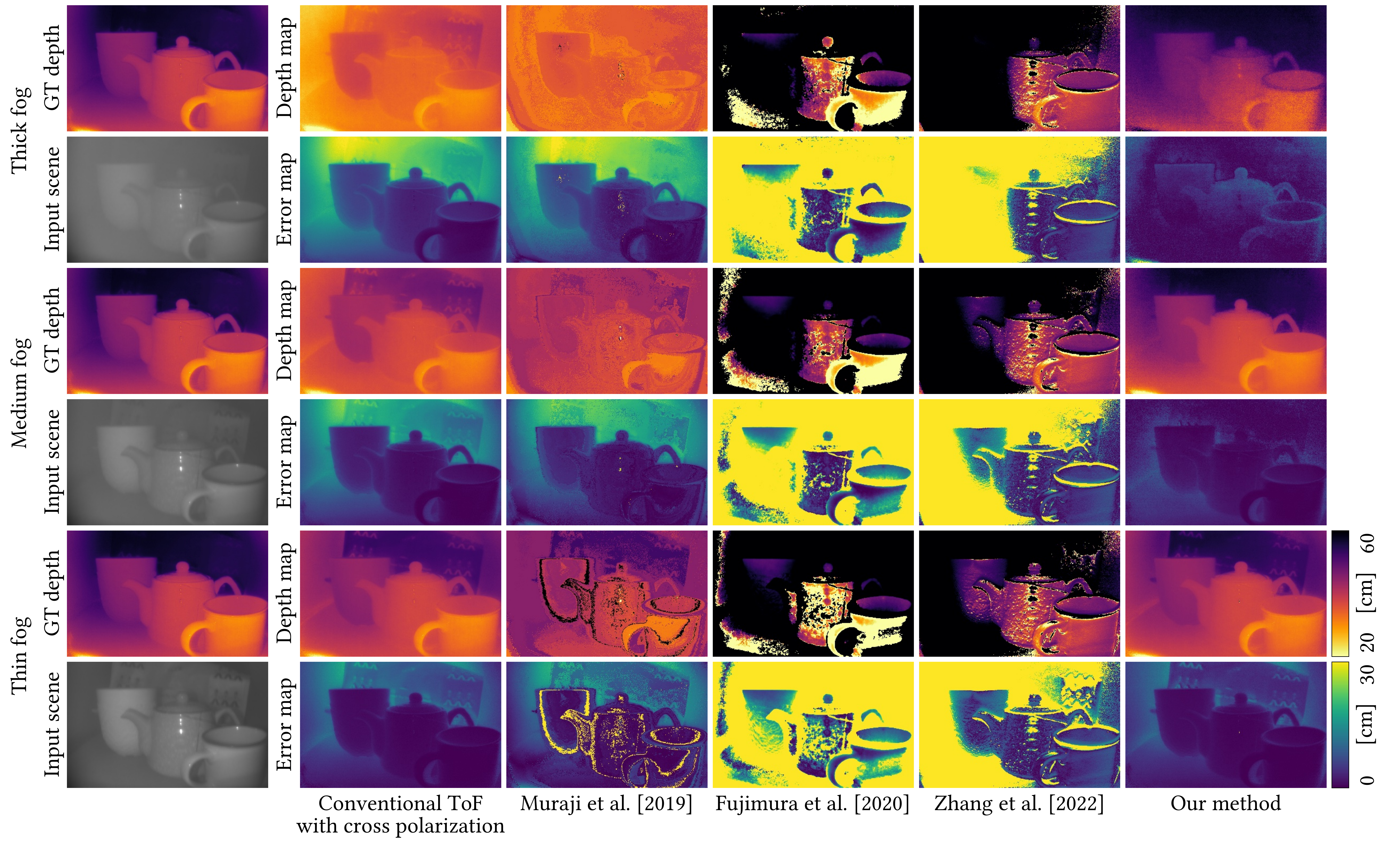}%
	\vspace{-5mm}%
	\caption[]{\label{fig:result27}%
    \emph{Scene: Cups.} We additionally compare our method to state-of-the-art ToF methods for scattering environments.
	}%
\end{figure*}

\end{document}